%% file: main.tex
\documentclass{applemlr}
\input{apple_preamble}

\newcommand*\methodname{{STARFlow-V}}
\newcommand*\ordering{Global–Local Ordering}

\NewDocumentCommand{\ying}{ mO{} }{\textcolor{teal}
{\textsuperscript{\textit{Ying}}\textsf{\textbf{\small[#1]}}}}
\title{STARFlow-V:~~End-to-End Video Generative Modeling with Normalizing Flows}

\author{Jiatao Gu}
\author{Ying Shen}
\author{Tianrong Chen}
\author{Laurent Dinh}
\author{Yuyang Wang}
\author{Miguel Ángel Bautista}
\author{David Berthelot} 
\author{Josh Susskind}
\author{Shuangfei Zhai}

\affiliation{Apple}
\abstract{
Normalizing flows (NFs) are end-to-end likelihood-based generative models for continuous data, and have recently regained attention with encouraging progress on image generation. Yet in the video generation domain, where spatiotemporal complexity and computational cost are substantially higher, state-of-the-art systems almost exclusively rely on diffusion-based models. 
    In this work, we revisit this design space by presenting STARFlow-V, a normalizing flow-based video generator with substantial benefits such as end-to-end learning, robust causal prediction, and native likelihood estimation.
    Building upon the recently proposed STARFlow, STARFlow-V operates in the spatiotemporal latent space with a global-local architecture which restricts causal dependencies to a global latent space while preserving rich local within-frame interactions. This eases error accumulation over time, a common pitfall of standard autoregressive diffusion model generation. 
    Additionally, we propose \emph{flow-score matching}, which equips the model with a light-weight causal denoiser to improve the video generation consistency in an autoregressive fashion. 
    To improve the sampling efficiency, STARFlow-V employs a video-aware Jacobi iteration scheme that recasts inner updates as parallelizable iterations without breaking causality. 
    Thanks to the invertible structure, the same model can natively support text-to-video, image-to-video as well as video-to-video generation tasks. Empirically, STARFlow-V achieves strong visual fidelity and temporal consistency with practical sampling throughput relative to diffusion-based baselines. 
    These results present the first evidence, to our knowledge, that NFs are capable of high-quality autoregressive video generation, establishing them as a promising research direction for building world models.
}

\metadata[Code]{\url{{https://github.com/apple/ml-starflow}}}
\metadata[Correspondence]{\sffamily Jiatao Gu (\url{jgu32@apple.com})}
\date{\sffamily\today}

\begin{document}

\maketitle
\applefootnote{\textcolor{textgray}{\sffamily Work done while
JG holding a joint affiliation with University of Pennsylvania, and 
YS working as a research intern at Apple MLR.}}

\input{secs/01-intro}

\input{secs/02-preli}

\input{secs/03-method}

\input{secs/04-exp}

\input{secs/04-results}

\input{secs/06-conclusion}

\bibliographystyle{plainnat}
\bibliography{iclr2026_conference}

%\newpage
\appendix

\input{secs/07-appx}

\applefootnote{ \textcolor{textgray}{\sffamily Apple and the Apple logo are trademarks of Apple Inc., registered in the U.S. and other countries and regions.}}

\end{document}

%% file: apple_preamble.tex
\usepackage{amsmath}
\usepackage{enumerate}
\usepackage{algorithm}
\usepackage{algpseudocode}
\usepackage{amsfonts}
\usepackage{amsthm}
\usepackage{cleveref}
\usepackage{diagbox}
\usepackage{colortbl}
\usepackage{amssymb}
\usepackage{xspace}
\usepackage{wrapfig}
\usepackage{adjustbox}
\usepackage{tabularx}
\usepackage{booktabs}
\usepackage{mathtools}
\usepackage{tikz}
\usepackage{enumitem}
\usepackage{silence}
\usepackage{dsfont}
\usepackage[table]{xcolor}
\usepackage[dvipsnames]{xcolor}
\usepackage{multirow}
\usepackage{makecell}
\usepackage{xfakebold}
\input{math_commands}

\definecolor{textgray}{HTML}{6E6E73}
\usetikzlibrary{positioning, calc}
\usetikzlibrary{decorations.pathmorphing}

\makeatletter
\patchcmd{\wrong@fontshape}{\@gobbletwo}{}{}{}
\makeatother
\numberwithin{equation}{section}
\makeatletter
\AtBeginDocument{
  \urlstyle{sf}
  
}
\makeatother

\definecolor{light}{RGB}{125, 125, 125}
\crefname{tcb@cnt@pbox}{code}{code}
\Crefname{tcb@cnt@pbox}{Code}{Code}
\crefname{assumption}{assumption}{assumption}
\Crefname{assumption}{Assumption}{Assumptions}

\newtcolorbox[auto counter]{pbox}[2][]{
  colback=white,
  title=Code~\thetcbcounter: #2,
  #1,fonttitle=\sffamily,
  fontupper=\sffamily,
  arc=2pt,
  colframe=bgcolor,
  coltitle=fgcolor,
  colbacktitle=bgcolor,
  toptitle=0.25cm,
  bottomtitle=0.125cm
}

\makeatletter
\newcommand\applefootnote[1]{%
  \begingroup
  \renewcommand\thefootnote{}%
  \renewcommand\@makefntext[1]{\noindent##1}%
  \footnote{#1}%
  \addtocounter{footnote}{-1}%
  \endgroup
}
\makeatother

\definecolor{cverbbg}{gray}{0.90}

%% file: math_commands.tex
\usepackage{amsmath,amsfonts,bm}

\def\eqref#1{equation~\ref{#1}}
\def\1{\bm{1}}

\def\vm{{\bm{m}}}

\def\vt{{\bm{t}}}
\def\vu{{\bm{u}}}

\def\vx{{\bm{x}}}
\def\vy{{\bm{y}}}
\def\vz{{\bm{z}}}

\DeclareMathAlphabet{\mathsfit}{\encodingdefault}{\sfdefault}{m}{sl}
\SetMathAlphabet{\mathsfit}{bold}{\encodingdefault}{\sfdefault}{bx}{n}

\newcommand{\pdata}{p_{\rm{data}}}

%% file: secs/01-intro.tex
\begin{figure}[!htbp]
    \centering
    \includegraphics[width=\textwidth]{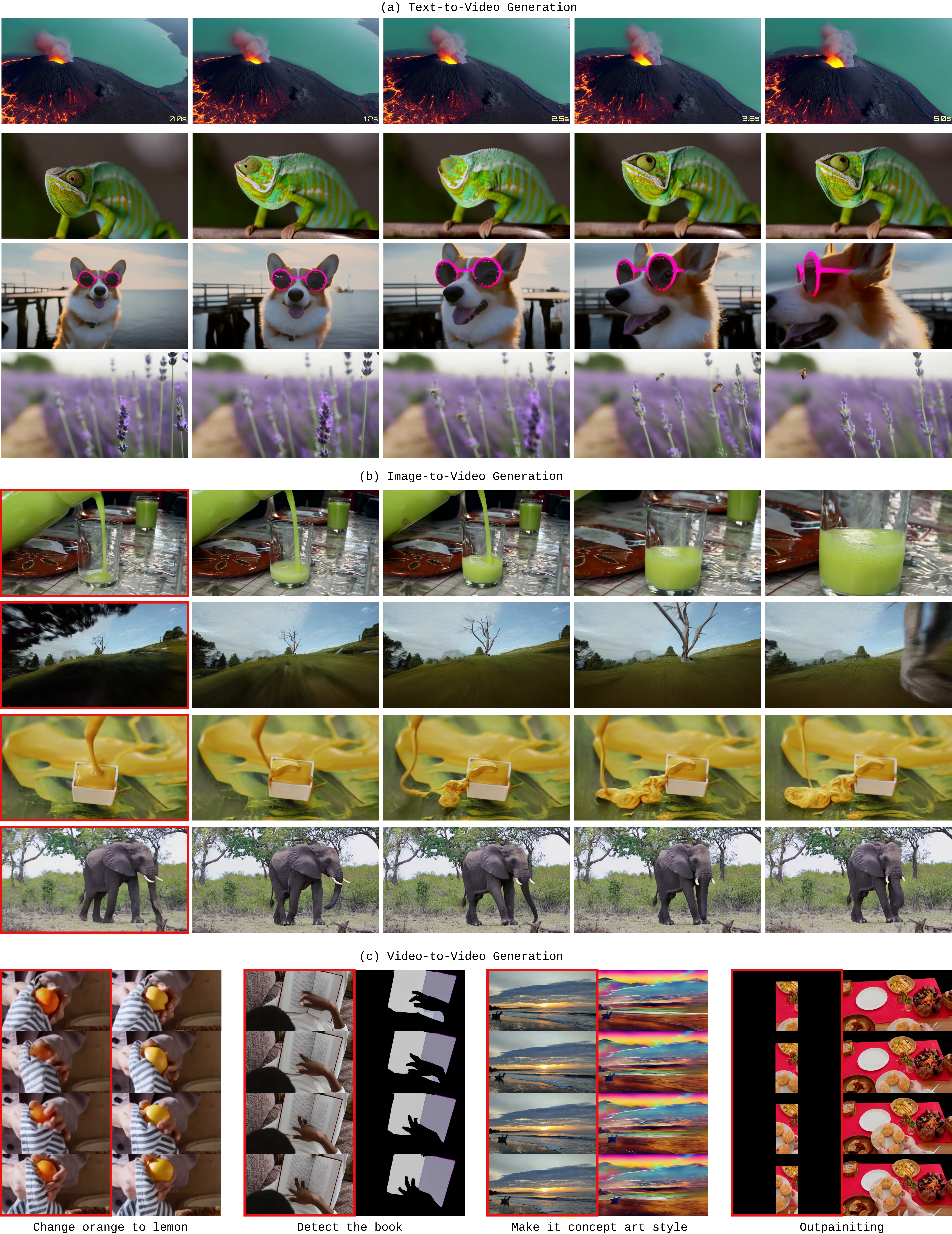}
    \caption{Samples from \methodname{} across three tasks. All videos are 480 p at 16 fps. Red boxes mark the conditioning inputs. The same autoregressive architecture is used for all tasks with no task-specific modifications. \textbf{Please find more generated videos and comparisons in the released code~\url{https://github.com/apple/ml-starflow}.}
    }
    \label{fig:teaser}
\end{figure}

\section{Introduction}
Deep generative modeling has advanced rapidly with breakthroughs across language~\citep{achiam2023gpt,openai2024gpt4o}, images~\citep{podell2023sdxl,batifol2025flux,wu2025qwen}, and videos~\citep{sora2024,wan2025wan,veo3_2025} domains. 
Among these modalities, \emph{video generation} is uniquely demanding: beyond high perceptual quality, models must capture rich spatiotemporal structure, remain robust over long horizons, and often operate under causal constraints for  streaming and interactive use.
Such capabilities are central not only to creative media~\citep{ye2025stylemaster,yuan2025identity}, but also to emerging \emph{world models} for gaming, simulation and embodied AI~\citep{ha2018worldmodels,yang2023learning,hu2023gaia,deepmind2024genie2,hafner2025dreamerv3}.

Recent scaling of data, model capacity, and compute has pushed video generation to new levels of fidelity~\citep{yangcogvideox,kong2024hunyuanvideo,kondratyuk2024videopoet,yu2024language,wan2025wan,seawead2025seaweed,gao2025seedance}.
In this space, \emph{diffusion-based} approaches~\citep{ho2020denoising,rombach2022high,peebles2023scalable,lipman2023flow,esser2024scaling} have emerged as the dominant backbone for text- and image-conditioned video synthesis, thanks to their strong empirical performance and flexible conditioning mechanisms.
Standard diffusion models are trained by corrupting frames with noise drawn from a schedule and learning a denoiser that inverts this process one step at a time, which leads to an iterative sampling procedure at inference.
For offline generation this formulation works well, but the parallel denoising of multiple frames is inherently non-causal: future frames can influence earlier ones, making it less natural to apply in streaming or interactive settings that require strictly causal rollouts.
Causally conditioned and sequential diffusion variants~\citep{chen2024diffusion,huang2025self} mitigate some of these issues, but still inherit the need to simulate noise at different timesteps and frames during training and can exhibit train--test mismatch during long-horizon autoregressive generation.

In parallel, \emph{normalizing flows} (NFs)~\citep{pmlr-v37-rezende15,dinh2014nice,dinh2016density} offer a distinct, likelihood-based alternative.
NFs are continuous end-to-end generative models that provide exact log-likelihood evaluation, non-iterative sampling, and native support for invertible feature mappings.
After an initial wave of work~\citep{dinh2016density,kingma2018glow}, they received relatively less attention compared to diffusion models, but have recently regained interest with encouraging progress on image generation~\citep{zhainormalizing,gu2025starflow,zheng2025farmer}.
In particular, STARFlow~\citep{gu2025starflow} shows that parameterizing an ``autoregressive normalizing flow'' with a Transformer and operating in a latent space allows flows to scale competitively in the high-resolution image domain.
Yet, in the video domain---where complexity and computational cost are substantially higher---state-of-the-art systems almost exclusively rely on diffusion, and it remains unclear whether NFs can be practical for video.

In this work, we revisit this design space and introduce \methodname{}, a normalizing-flow-based video generator that combines end-to-end training with causal, likelihood-based modeling. 
Building on STARFlow~\citep{gu2025starflow}, \methodname{} operates in a spatiotemporal latent space with a \emph{global--local} architecture: a compact global latent sequence carries long-range temporal context, while local latent blocks preserve fine-grained within-frame structure. 
By delegating temporal reasoning to this high-level space, the model mitigates the accumulation of autoregressive errors that commonly plagues diffusion-based video generators. 
As observed in TARFlow~\citep{zhai2024normalizing}, training flows on slightly perturbed data with a subsequent denoising step can significantly improve robustness. Unlike existing methods~\citep{zhai2024normalizing,gu2025starflow}, we propose \emph{flow-score matching}, which learns a lightweight causal denoiser to enhance temporal consistency in video scenarios. 
To further improve efficiency, \methodname{} employs a video-aware Jacobi-style update scheme that recasts inner refinement steps as parallelizable iterations. 
Finally, owing to its invertible nature, the same backbone naturally supports text-to-video (T2V), image-to-video (I2V), and video-to-video (V2V) generation by simply changing the form of the conditioning signal.

Across all benchmarks, \methodname{} attains visually coherent and temporally stable generations while maintaining practical sampling speed relative to diffusion-based models. We believe this provides \textbf{initial} evidence that NFs are capable of high-quality autoregressive video generation and potentially world models.

%% file: secs/02-preli.tex
\section{Background}

\subsection{Video Generative Models}
Given $N$ frames $\vx_{1:N}=(\vx_1,\ldots,\vx_N)$ and optional conditioning $C$ (\textit{e.g.}, text, image, audio, layout, camera), video generative models seek to model the joint distribution of all frames $p(\vx_{1:N}\mid C)$ and sample novel videos from the learned model. 
While earlier work explored GANs~\citep{Vondrick2016VGAN,Tulyakov2018MoCoGAN,Skorokhodov2022StyleGANV}, VAEs~\citep{Babaeizadeh2018SV2P,castrejon2019improved,Wu2021GHVAE}, and discrete autoregressive models~\citep{Yan2021VideoGPT,yu2024language,kondratyuk2024videopoet}, the field has largely converged on diffusion-based methods~\cite{ho2022video,ho2022imagen}. Spurred by the release of Sora~\citep{videoworldsimulators2024}, DiT-style approaches~\citep{peebles2023scalable} have shown strong generalization at scale~\citep{gao2025seedance,wan2025wan,veo3_2025}. A key distinction from prior paradigms is that training of diffusion-based models is \textit{Not End-to-End}: diffusion-based models corrupt frames with noise at randomly sampled levels and train a denoiser to invert this process, optimizing an objective closely related to the lower bound of $\log p(\vx_{1:N}\mid C)$. 
This setup incurs high cost—especially for video—as each update supervises only a single noise level.
At inference time, one sample is generated by iteratively denoising from Gaussian noise. 

%\noindent\textbf{Autoregressive Video Synthesis}~
Diffusion-based video generation is typically non-causal: all frames are corrupted with noise and denoised in parallel~\citep{ho2022video}. Yet many real-world applications demand causal, often interactive synthesis (\emph{e.g.}, online streaming, video games, robotics), where frames must be produced sequentially.
% Autoregressive (AR) models have been widely adopted for video generation due to their ability to factorize high-dimensional distributions into tractable conditionals.
% Given a sequence of $N$ video frames $\vx_{1:N} = (\vx_1, \vx_2, \ldots, \vx_N)$, the joint distribution is factorized via the chain rule as:
% \begin{equation}
%     p(\vx_{1:N} \mid C) = \prod_{n=1}^N p(\vx_n \mid \vx_{<n}, C),
% \end{equation}
% where each conditional distribution $p(\vx_n \mid \vx_{<n}, C)$ generates the $n$-th frame conditioned on all previous frames and the additional condition contexts $C$ (\textit{e.g.}, label, text, image, and \textit{etc}.).  
% There exist two typical classes of AR video generation approaches:
% (1) Discrete autoregressive generation~\citep{hongcogvideo,kondratyuk2024videopoet,yu2024language}.
% These methods generate perform causal discrete token prediction within video frame sequence similar to large language models. However, these approaches are constrained with the information loss due to discretization.
% %Given a sequence of $D$ tokens $\mathbf{u}_{1:D}$ representing a video, the joint distribution is factorized as:
%\begin{equation}
%    p(\vu_{1:D} \mid C) = \prod_{d=1}^D p(\vu_d \mid \vu_{<d}, C).
%\end{equation}
Autoregressive (AR) diffusion models~\citep{chen2024diffusion,song2025history,yin2025slow}—a line of work that combines chain-rule factorization with diffusion—aim to alleviate prior limitations by introducing asynchronous, frame-wise noise schedules during training, modeling each conditional \(p(\vx_n \mid \vx_{<n})\) as a diffusion process. Despite their strengths, AR generation typically suffers from \emph{exposure bias}: during training, models condition on ground-truth contexts, whereas at inference they must rely on their own (imperfect) predictions. This train–test mismatch compounds over time, degrading long-horizon video quality. The \emph{non–end-to-end} nature of diffusion training further exacerbates this gap, though recent efforts such as Self-Forcing~\citep{huang2025self} seek to mitigate it via sequential post-training with distillation objectives. However, they are not readily applicable in the pre-training stage on raw video data.
% \paragraph{Video Latent Space}
% Directly modeling long-duration and high-resolution videos in pixel space is computationally challenging. 
% Therefore, recent models typically operate in a compressed latent space~\citep{rombach2022high}.
% In particular, video frames are encoded with a 3D causal variational autoencoder (VAE)~\citep{yangcogvideox,wan2025wan}, compressing both spatial and temporal dimensions while enforcing causality along the temporal axis. Throughout, we adopt latent-space representations unless explicitly indicated otherwise. 
%In particular, given input video frames $\vx_{1:N} \in \R^{N \times H \times W \times C}$, the encoder $\mathcal{E}$ produces a sequence of spatio-temporal latents  $\tilde{\vx}_{1:D} = \mathcal{E}(\vx_{1:N}) \in \R^{D \times d_h}$, where $d_h$ is the latent channel dimension and $D = \frac{N}{r_t} \times \frac{H}{r_h} \times \frac{W}{r_w}$, with $r_t, r_h, r_w$ denoting the temporal and spatial downsampling factors of the encoder.

\subsection{Autoregressive Normalizing Flows}

Normalizing flows~\citep[NFs;][]{pmlr-v37-rezende15,dinh2014nice,dinh2016density,kingma2018glow,ho2019flow++} are likelihood-based generative models built from invertible transformations.
Given a continuous input $\vx\!\sim\!\pdata$, $\vx\in\mathbb{R}^D$, an NF learns a bijection $f_\theta:\mathbb{R}^D\!\to\!\mathbb{R}^D$ that maps data $\vx$ to latents $\vz=f_\theta(\vx)$.
Unlike diffusion models, NFs are trained \emph{end-to-end} via a tractable maximum-likelihood objective derived from the change-of-variables formula:
\begin{align}
\label{eq:nf_loss}
\mathcal{L}_{\text{NF}}(\theta) %&= ~\mathbb{E}_{\vx\sim\pdata}\!\left[\log p_{\text{NF}}(\vx;\theta)\right] \nonumber \\
= \mathbb{E}_{\vx}\!\left[\log p_{0}\!\big(f_\theta(\vx)\big)
+ \log\!\left|\det\!\left(J_{f_\theta}(\vx)\right)\right|\right],
\end{align}
where the first term encourages mapping data to high-density regions of a simple prior $p_0$ (e.g., standard Gaussian), and the Jacobian term $J_{f}$  accounts for the local volume change induced by $f_\theta$, preventing collapse.
Once trained, sampling is immediate via inversion: draw $\vz\!\sim\!p_0(\vz)$ and set $\vx=f_\theta^{-1}(\vz)$.
Historically, however, NFs have been viewed as less competitive than diffusion models due to architectural rigidity and training instability~\citep{dinh2016density}.

Recently, TARFlow~\citep{zhainormalizing} and its scalable extension, STARFlow~\citep{gu2025starflow}, have revisited normalizing flows as next-generation backbones for generative modeling. Both methods instantiate autoregressive flows (AFs)~\citep{kingma2016improved,papamakarios2017masked}—NFs whose invertible transformations are parameterized autoregressively—and use causal Transformer blocks, in the style of LLMs, as their primary building units.
Formally, STARFlow~\citep{gu2025starflow} stacks \(T\) autoregressive flow blocks with alternating directions,
%Let \(\vx^0\!=\!\vx\). The overall flow is $\vz = f(\vx) := (f^T_\theta \circ f^{2}_\theta \circ \cdots \circ f^1_\theta)(\vx)$, 
where each block applies an affine transform whose parameters are predicted by a causal Transformer under a (self-exclusive) causal mask \(\vm\):
\begin{align}
    \vz
\;=\; 
\left[\vx - \mu_\theta \big(\vx \odot \vm\big)\right] / 
     {\sigma_\theta \big(\vx \odot \vm\big)},
\;
\sigma_\theta(\cdot) > 0,
\label{eq.affine_f}
\end{align}
where $\vx, \vz$ are the input and output of each block, \(\odot\) denotes the Hadamard product. As shown in STARFlow~\citep{gu2025starflow}, \(T\!\ge\!3\) blocks suffice for universal density modeling where masks alternate between left-to-right (\(\rightarrow\)) and right-to-left (\(\leftarrow\)) to capture bidirectional dependencies.
% (Equivalently, one may parameterize $\sigma_\theta=\exp s_\theta$ to enforce positivity.)
% Training is still performed end-to-end:
% \begin{equation}
%     \max_\theta~ \mathbb{E}_{\vx\sim\pdata} \log \paf(\vx;\theta) = -\frac{1}{2}\|\vz\|^2_2 - \sum_{t=1}^T\sum_{d=1}^{D} \log \sigma^t_\theta(\vx^{t}_{\pi_{<d}}),
%     \label{eq:tarflow_loss}
% \end{equation}
% where $\vx^{t} = f^t_\theta (\vx^{t-1})$ defines the forward propagation (\Cref{eq.af_step}).

Despite STARFlow demonstrating competitive  quality with state-of-the-art diffusion~\citep{podell2023sdxl,esser2024scaling} on large-scale text-to-image tasks, evidence for normalizing flows in video generation remains sparse. To our best knowledge, the only prior NF-based video model is VideoFlow~\citep{kumar2019videoflow}, which builds on Glow~\citep{kingma2018glow}  and is constrained by limited capacity, low resolution, and domain-specific settings. Compared to images, video generation is substantially more challenging for NFs due to higher spatiotemporal dimensionality. Nevertheless, we argue that normalizing flows—exemplified by STARFlow—are a natural fit for video modeling, especially in autoregressive settings.

% Autoregressive flows~\citep[AFs,][]{kingma2016improved,papamakarios2017masked} are an important class of NFs, where the invertible transformation is constructed by stacking multiple autoregressive transformations with alternating directions.  
% In their simplest affine form, AFs define the forward $(\vx \to \vz)$ and inverse $(\vz \to \vx)$ transformations as:
% \begin{equation}
%      \vz_d  = \left(\vx_d - \mu_\theta(\vx_{<d})\right) / \sigma_\theta(\vx_{<d}), ~~~~~~
%      \vx_d = \mu_\theta(\vx_{<d}) + \sigma_\theta(\vx_{<d}) \cdot \vz_d,~~  \forall d \in [1, D],
%      \label{eq.af_step}
% \end{equation}
% where $\mu_\theta$ and $\sigma_\theta$ are learnable functions of the preceding variables $\vx_{<d}$. 
% This affine parameterization ensures invertibility and yields a triangular Jacobian, making the log-determinant efficient to compute while preserving exact likelihood training.  

% Recently, TARFlow~\citep{zhainormalizing} has revisited AFs and further extended them with powerful Transformer architectures.
% Unlike classical AFs that operate on a per-dimension basis with masked MLPs, 
% TARFlow adopts block-wise autoregression, predicting a block of affine transformation parameters at a time through masked Transformer layers.  
% This design enables the model to capture longer-range dependencies while scaling more effectively to high-dimensional data.  

% For an input presented in the form of a sequence $\vx \in \R^{N \times D}$, where $N$ is the sequence length and $D$ is the dimension of input data.

%% file: secs/03-method.tex
\section{\methodname{}}
We propose \methodname{}, a novel paradigm for video generation based on normalizing flows. 
While inspired by STARFlow~\citep{gu2025starflow}, \methodname{} is not a direct port to the video domain; it introduces several architectural redesigns and algorithmic innovations tailored to spatiotemporal data.
In what follows, we present the architecture and its autoregressive formulation (\Cref{sec:ar_video}), the training procedure (\Cref{sec: training}), the inference pipeline (\Cref{sec: inference}), and applications enabled by our model (\Cref{sec:application}).

% Given a sequence of $N$ video frames $\vx^{1:N} = (\vx^1, \vx^2, \ldots, \vx^N)$, 
% where each frame $\vx^n \in \mathbb{R}^{H \times W \times C}$ is an image with resolution $H \times W$ and $C$ channels, \methodname{} aims to learn a generative model over the joint distribution $p_\theta(\vx^{1:N})$. An image can be viewed as a special case where $N=1$. 
% Specifically, \methodname{} is trained jointly on images and videos within the same framework.
% Following prior AR video generation approaches~\citep{yangcogvideox,wan2025wan}, \methodname{} operates in a compressed latent space to efficiently handle long-duration, high-resolution videos. 
% Specifically, \methodname{} first encodes $N$ video frames $\vx_{1:N}$ into compressed latent representation $\tilde{\vx}_{1:D} = \mathcal{E}(\vx_{1:N})$ using a 3D causal variational autoencoder (VAE)~\citep{wan2025wan}.
% The autoregressive flow model operates directly on $\tilde{\vx}_{1:D}$, while generated samples are later decoded back to pixel space via the VAE decoder $\mathcal{D}(\tilde{\vx}_{1:D})$.  
% This significantly reduces dimensionality while preserving semantic and motion information, and ensures scalability to long videos.

\subsection{Proposed Model}
\label{sec:ar_video}
For a video \(\vx \in \mathbb{R}^{N\times H\times W\times D}\),
each frame \(\vx_n\) is flattened to \(\mathbb{R}^{HW\times D}\),
\(\vx_n = (\vx_{n,1}, \ldots, \vx_{n,HW})\), and all frames are concatenated
into a sequence of \(NHW\) tokens.
We operate in a compressed latent space using a pretrained 3D causal
VAE~\citep{wan2025wan}.
\methodname{} models the joint distribution \(p_\theta(\vx)\) via an
invertible mapping \(f_\theta\) implemented as autoregressive flows
(\Cref{eq.affine_f}).
Following \citet{gu2025starflow}, we use a \emph{deep–shallow}
decomposition \(f_\theta = f_D \circ f_S\), where a small stack of
\emph{shallow} flow blocks with alternating (left-to-right / right-to-left)
masks maps \(\vx\) to intermediate latents \(\vu = f_S(\vx)\), and a
\emph{deep} causal-Transformer flow \(f_D\) then maps \(\vu\) to the prior,
producing \(\vz = f_D(\vu)\).
By the change-of-variables formula,
\begin{equation}
p_\theta(\vx) \;=\; p_0(\vz)\,
\bigl|\det J_{f_D}(\vu)\bigr|\,
\bigl|\det J_{f_S}(\vx)\bigr| ,
\label{eq.starflow_v}
\end{equation}
where \(p_0\) is a simple prior (e.g., standard Gaussian).
Most capacity is allocated to the deep block \(f_D\) for semantic modeling,
while the shallow stack \(f_S\) handles local reshaping.
For videos, we can simply
treat all frames as one long token sequence: \(f_D\) follows a left-to-right
causal order over the video (causal across frames, raster order within each
frame), and \(f_S\) retains the alternating masks defined above.
Because \(f_S\) propagates information from future frames to past ones, this
naïve design yields a \emph{non-causal} video generator, motivating the
global–local restructuring described next.

\noindent\textbf{Global–Local Architecture}~
Observing that \(f_D\) is inherently autoregressive and that \(f_S\) mainly provides 
local refinements, we adapt the design into a \emph{global–local} structure: 
\(f_S\) is restricted to operate within each frame, while only \(f_D\) propagates 
global video context in a causal manner. More specifically, \Cref{eq.starflow_v} can be re-expressed as an autoregressive 
factorization over frames \(\vx_n\):
\begin{equation}
p_\theta(\vx)
= \prod_{n=1}^N p_\theta(\vx_n \mid \vx_{<n})
= %\underbrace{
\prod_{n=1}^N p_D(\vu_n \mid \vu_{<n})
%}_{\footnotesize p_0(\vz)\, \bigl|\det J_{f_D}(\vu)\bigr|}
   \bigl|\det J_{f_S}(\vx_n)\bigr| ,
    \label{eq.factorize}
\end{equation}
where \(\vu_n = f_S(\vx_n)\) denotes the local latents for frame \(\vx_n\). 
Here, the deep block is itself an autoregressive flow,
capturing both intra-frame raster ordering and inter-frame causal dependencies. 
% Inference proceeds by (i) inverting the deep flow $f_D$ to obtain \(\vu\) via left-to-right autoregressive sampling, and (ii) reconstructing frames independently through the shallow inverse, \(\vx_n = f_S^{-1}(\vu_n)\).

Formulating \methodname{} in a \emph{global–local} manner (\Cref{eq.factorize}) yields  several benefits:
\begin{enumerate}[leftmargin=1.5em,label=(\alph*)]
\item \textbf{Universality.}
\Cref{eq.factorize} preserves the universal approximation guarantee of STARFlow~\citep{gu2025starflow}:
the local stack \(f_S\) still realizes per-pixel infinite Gaussian mixtures via alternating causal masks,
so expressivity is not curtailed by restricting \(f_S\) to within-frame contexts.

\item \textbf{Robustness.}
Intuitively, \Cref{eq.factorize} can be viewed as a \textbf{continuous language model for videos}:
the deep-flow term \(p_D(\vu_n \mid \vu_{<n})\) acts as \emph{Gaussian Next-Token Prediction}
(cf.\ the affine form in \Cref{eq.affine_f}) in latent space, while the shallow flow supplies
the Jacobian factor \(\lvert \det J_{f_S}(\vx_n) \rvert\), yielding a flexible density over \(\vx\).
Compared to modeling \(\vx\) directly (arbitrarily multimodal), the latent \(\vu\) is unimodal at each step,
easier to regress, and more tolerant to small prediction errors. Crucially, the sampling phase
via \(f_D^{-1}\) conditions on previously generated \emph{latents} rather than pixels, so data-space errors
do not propagate forward, mitigating the compounding error typical of autoregressive diffusion.
Unlike diffusion-style noise conditioning~\citep{ho2022cascaded,chen2024diffusion}, which compromises
information to gain robustness and introduces extra parameters, our mappings \(\vu \leftrightarrow \vx\)
are invertible, avoiding information loss by construction.
%The deep distribution \(p_D(\vu)\) (see \Cref{eq.factorize}) can be viewed as an \emph{Autoregressive Gaussian} over latents (cf.\ the affine form in \Cref{eq.affine_f} under a causal mask \(\vm\)):
% \[
% p_D(\vu) \;\propto\; \mathcal{N}\!\big(\vu;\,\mu_D(\vu\odot\vm),\,\sigma_D^2(\vu\odot\vm)\big), 
% \]
%where $\vm$ is the causal mask.

\item \textbf{End-to-End Training.}
The whole model is still NF. Consequently, all parameters are trained by exact MLE via the change-of-variables objective—no per-step denoising schedule or surrogate loss—simplifying optimization and reducing train–test mismatch.

\item \textbf{Streamable Generation.}
At inference time, \(f_D^{-1}\) samples \(\vu_n\) causally (token-by-token, frame-by-frame), and \(f_S^{-1}\) decodes each frame independently given $\vu_n$. This process enables causal video synthesis since later frames cannot influence earlier ones.
\end{enumerate}

\begin{figure*}[t]
    \centering
    \includegraphics[width=0.95\linewidth]{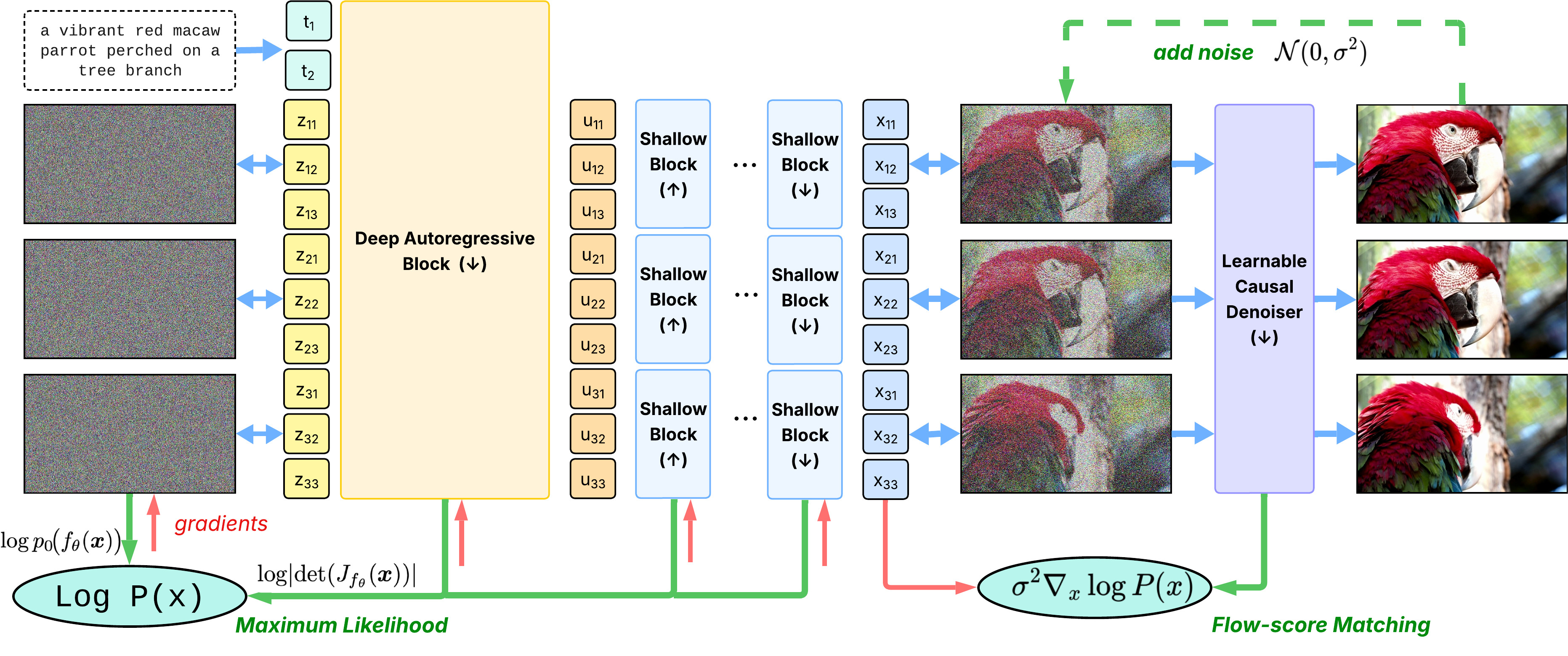}
    \caption{An illustrated pipeline of \methodname{} which shows (1) the proposed global-local architecture; (2) joint training with the learnable denoiser with the proposed Flow-score Matching.
    During sampling, \methodname{} takes the encoded text condition $\vt$ and transforms the noise $\vz$ through deep global block to intermediate features $\vu$, followed by several local shallow blocks to produce a slightly noised video. Finally, a learnable causal denoiser refines this output into the final clean video $\vx$. 
    }
    \label{fig:pipeline}
\end{figure*}

\subsection{Revisiting Noise-Augmented Training}
\label{sec: training}
As observed by \citet{zhai2024normalizing}, injecting \emph{small} noise into the data is crucial for stabilizing NF training. 
Concretely, we learn \methodname{} on a $\sigma$-smoothed density \(q_\sigma(\tilde{\vx}) = (p * \mathcal{N}(0, \sigma^2 I ))(\tilde{\vx})\). %with \(\tilde{\vx}=\vx+\sigma \cdot \epsilon\), \(\epsilon\sim\mathcal{N}(0, I)\). 
A side effect is that the model naturally generates slightly noisy samples, necessitating a post-processing step to recover the clean ones. We first examined the existing options for this purpose:
\begin{enumerate}[leftmargin=1.5em,label=(\alph*)]
\item \textbf{Decoder Fine-tuning}~
We followed STARFlow~\citep{gu2025starflow}, adopting their strategy of 
fine-tuning the VAE decoder to denoise noisy latents using a GAN objective~\citep{rombach2022high}. 
However, our preliminary experiments suggest that this approach is not readily applicable to 
\emph{3D causal} VAEs: under Gaussian-noised latent inputs, the decoder fails to maintain temporal consistency in the generated videos due to limited receptive fields.

\item \textbf{Score-based Denoising}~
Instead of decoder fine-tuning, TARFlow~\citep{zhai2024normalizing} proposes to denoise using the 
\emph{learned flow} itself via score-based updates. For a noisy sample $\tilde{\vx}\sim q_\sigma$, the continuity equation gives ${\partial_{\sigma}  \tilde{\vx}}= -\sigma \nabla_{\tilde{\vx}}\log q_\sigma(\tilde{\vx})$.
So for sufficiently small \(\sigma\), a single Euler step yields the Tweedie estimator:
\begin{equation}
\vx \;\approx\; \tilde{\vx} - \sigma\,\partial_{\sigma}\tilde{\vx} 
\;=\; \tilde{\vx} + \sigma^2 \nabla_{\tilde{\vx}}\log q_\sigma(\tilde{\vx}).
\label{eq.score_denoise}
\end{equation}
With normalizing flows, we replace \(q_\sigma\) by the learned density \(p_\theta\), and compute 
\(\nabla_{\tilde{\vx}}\log p_\theta(\tilde{\vx})\) via automatic differentiation through the flow, 
which amounts to an additional forward–backward pass.
However, this score-based denoising presents two issues:
\textbf{(1) Noisy gradients.} The learned density \(p_\theta\) is imperfect; its score 
\(\nabla_{\tilde{\vx}}\log p_\theta(\tilde{\vx})\) often contains high-frequency noise, which manifests as 
bright speckle-like artifacts—especially in regions with large motion; \textbf{(2) Non-causality of the score.} Even if \(p_\theta\) is modeled causally, the score 
\(\nabla_{\tilde{\vx}}\log p_\theta(\tilde{\vx})\) is, by definition, global: the gradient at time \(n\) depends 
on likelihood terms involving future frames \(m>n\). This breaks causality, undermining 
the promised streamable generation.
\end{enumerate}

\noindent\textbf{Proposed Approach: Flow-Score Matching}
To address these issues, we introduce a lightweight neural denoiser \(s_\phi\) trained alongside the flow \(f_\theta\) to regress the model’s score:
\begin{equation}
\label{eq:fsm_loss}
\mathcal{L}_{\text{denoise}}(\phi)
= \mathbb{E}_{\vx,\,\boldsymbol{\epsilon}}
\big\|\, s_\phi(\tilde{\vx}) \;-\; \sigma\,\nabla_{\tilde{\vx}}\log p_\theta(\tilde{\vx}) \,\big\|_2^2,
\qquad
\tilde{\vx}=\vx+\boldsymbol{\epsilon},\;\boldsymbol{\epsilon}\sim\mathcal{N}(0,\sigma^2 I).
\end{equation}
At inference, we replace the raw score in the update (cf.\ \Cref{eq.score_denoise}) with the learned denoiser \(s_\phi\).
This \emph{flow-score matching} (FSM) is simple yet effective. First, the smooth inductive bias of neural networks
suppresses stochastic high-frequency artifacts in \(\nabla_{\tilde{\vx}}\log p_\theta\). Second, we can encode causality directly in \(s_\phi\), re-ensuring streamable behavior.
Concretely, we parameterize \(s_\phi\) with a one–frame look-ahead while remaining globally causal (one-step latency)\footnote{Strictly causal (\(\le n\)) fails as temporal \emph{differences} are pivotal to determining the denoising direction.}. We approximate the score at step \(n\) by
%\begin{equation}
%\label{eq:fsm_local}
$s_\phi(\tilde{\vx}_{\le n+1}) \;\approx\; \bigl(\sigma\,\nabla_{\tilde{\vx}}\log p_\theta(\tilde{\vx})\bigr)_n.$
%\end{equation}
Finally, we train \(s_\phi\) jointly with \(f_\theta\) at \textbf{minimal overhead}: 
since \(f_\theta\) is trained by maximizing \(\log p_\theta\), we cache the 
input gradients from the backward pass and reuse it as the target for \(s_\phi\).

% However, directly applying the score-based denoising to \methodname{} often introduces visible artifacts in the generated videos. 
% To address this, \methodname{} introduces a causal learnable self-denoiser that explicitly predicts the score of the noisy latent distribution. 
% The learnable self-denoiser $s_\phi(\cdot)$ is trained by minimizing

% \begin{equation}
%     \mathbb{E}_{\vy\sim \mathcal{N}(\vx; \sigma^2I), \sigma}\left[-\sigma \log p_\theta(\vy) + \|s_\phi(\vy, \sigma) - \sigma \left(\nabla_{\vy}\log p_\theta(\vy)\right)_{\text{stopgrad}} \|^2_2\right]
% \end{equation}

\begin{comment}
Following the noised augmented training strategy introduced in TARFlow~\citep{zhainormalizing}, \methodname{} is also trained on noisy distribution $q(\vy)$ where $\vy = \tilde{\vx} + \epsilon$ for $\epsilon \sim \mathcal{N}(0, \sigma^2 I)$. 
This augmentation improves robustness but has the side effect that the model 
naturally generates noisy samples rather than clean ones.  

As a remedy, TARFlow proposes to denoise the generated samples through training-free score-based denoising:
\begin{equation}
    \vz \sim p_0, \vy := f_\theta^{-1}(\vz), \tilde{\vx} := \vy + \sigma^2 \nabla_\vy \log p_\theta(\vy),
\end{equation}
where $\nabla_\vy \log p_\theta(\vy)$ is the score function estimated by the generative model itself. This allows the noisy $\vy$ to be corrected into a cleaner $\tilde{\vx}$ without additional training.  
\end{comment}

\input{tabs/vbench}

\subsection{Fast Inference}
\label{sec: inference}

While \methodname{} leverages parallel computation during training via causal masking, 
generation at inference time is carried out sequentially (one token at a time) through multiple AF blocks, 
which can be \emph{extremely} computationally demanding for long video sequences.
For instance, generating a $5$s 480p video under $16$ fps using a pre-trained 3B parameter model requires over $30$ minutes, which is far from real-time performance.
To enable fast inference, we introduce two strategies:

\noindent\textbf{Block-wise Jacobi Iteration} Rather than sampling continuous tokens strictly autoregressively, we accelerate inference by
recasting inversion as solving a nonlinear fixed-point system with parallel solvers such as
Jacobi iteration~\citep{porsching1969jacobi,kelley1995iterative}, a strategy recently used to
speed up autoregressive models~\citep{song2021accelerating,teng2024accelerating,liu2025accelerate,zhang2025inference}.
Specifically, the inverse of \Cref{eq.affine_f} can be written as the fixed-point equation
\begin{equation}
\label{eq:affine_inv_fp}
\vx \;=\; \mu_\theta(\vx \odot \vm)\;+\; \sigma_\theta(\vx \odot \vm)\cdot \vz,
\end{equation}
where \(\vm\) is a (self-exclusive) causal mask. This induces a \emph{triangular} system that admits convergence under nonlinear Jacobi iteration~\citep{saad2003iterative}: starting from an initial sequence estimate \(\vx^{(0)}\), iterate $\vx^{(k+1)} = \mu_\theta(\vx^{(k)} \odot \vm) + \sigma_\theta(\vx^{(k)} \odot \vm )\cdot \vz$ until a converge criterion is satisfied. We monitor a scale-normalized residual, \(
{\|\vx^{(k+1)} - \vx^{(k)}\|_2^2}/{\|\vx^{(k+1)}\|^2_2} < \tau
\) with $\tau = 0.001$ by default. 
Although the worst-case iteration count scales with sequence length (\emph{e.g.}, near-Markovian process), video generation exhibits strong global structure, substantially accelerating convergence in practice. The procedure is also \emph{guidance-compatible}, as proposed in ~\citep{gu2025starflow}, which involves computing the guided parameters \(\hat{\mu}\) and \(\hat{\sigma}\) and then substituting them.

To further accelerate sampling, we adopt a block-wise Jacobi scheme
in the spirit of \citet{song2021accelerating,liu2025accelerate}. 
The token sequence is partitioned into contiguous blocks of size \(B\), which are processed sequentially across blocks but in parallel within each block. 
Within each block we run the Jacobi updates,
while states from completed blocks are cached as context (\emph{e.g.}, KV cache) for subsequent blocks—analogous to standard AR
inference. We also apply a video-aware initialization: for a new frame, the initial estimate \(\vx^{(0)}_{n+1}\) is initialized from the previously converged frame \(\vx^{(k)}_{n}\).
Overall, we adopt block-based iteration within each AF block, yielding $\approx \textbf{15}\times$ lower inference latency relative to standard autoregressive decoding, while preserving visual fidelity.

\noindent\textbf{Pipelined Decoding}
As described in \Cref{sec:ar_video}, the global–local design applies standard global left-to-right autoregression in the deep block $f_D$, while the shallow blocks $f_S$ traverse each frame independently. This enables a pipelined schedule (analogous to pipeline parallelism~\citep{huang2019gpipe}): $f_D$ runs continuously without waiting on $f_S$, and, in parallel, $f_S$ threads consume $f_D$’s outputs, immediately refine them, and then denoise. Because $f_D$ is typically the slowest stage, end-to-end latency is dominated by the deep block.

\subsection{Versatility Across Tasks}
\label{sec:application}
\methodname{} can be trained for different video generation tasks. By default, \methodname{} is trained for text-to-video generation on large-scale text–video pairs. Without modifying the backbone, we support the following settings:
\begin{enumerate}[leftmargin=1.5em,label=(\alph*)]
\item \textbf{Image-to-Video Generation.}
We directly treat the first frame as observed conditioning. Owing to the invertiblity, \emph{no separate encoder is required}: we encode the observed frame via the flow forward to initialize the KV cache; subsequent frames are then generated.

\item \textbf{Video-to-Video Generation.}
Given a source clip $\vx_{0:T}$, we treat all frames as observed conditioning and—thanks to invertibility—use the same backbone to flow-encode them and populate the KV cache. The model then autoregressively rolls out the target clip $\hat{\vx}_{0:T}$ under optional task cues (e.g., in/outpainting masks, edit text, camera/pose), copying through unedited regions while synthesizing edits. This mirrors our image-to-video path but operates framewise over the whole clip without a separate encoder.

\item \textbf{Longer Generation.}
Our model generates videos far longer than those seen during training via a sliding-window (chunk-to-chunk) schedule in the deep block. After producing a latent chunk $\vu$, we warm-start the next step by rebuilding the KV cache: we re-run $f_D$ on the last $\Delta$ latents (the overlap) and then continue autoregression to synthesize the next $N\!-\!\Delta$ latents. $f_S$ then process the latents per frame, enabling streaming output. To mitigate boundary mismatch, we randomly drop the first frame during training to simulate restart.

%\item \textbf{Controllable Video Generation.}
%For non-video conditioned synthesis (e.g., camera-controlled and force-guided generation), we freeze the backbone and train a control module that injects control tokens into the autoregressive steps. This enables precise control without altering the core architecture.
\end{enumerate}

%% file: tabs/vbench.tex
\begin{table*}[t]
\centering
\resizebox{\textwidth}{!}{
\begin{tabular}{lccccccccc}
\toprule
Model  & Total & Quality & Semantic & Aesthetic & Object & Multi Obj. & Human & Spatial & Scene \\
\midrule
\rowcolor{gray!15} \multicolumn{10}{l}{\textit{Closed-source models}} \\
%Gen-2~\citep{germanidis2023gen2}          & 80.58 & 82.47 & 73.03 & 66.96 & 90.92 & 55.47 & 89.20 & 66.91 & 48.91 \\
% Kling (2024-07) & -   & 81.85 & 83.39 & 75.68 & 61.21 & 87.24 & 68.05 & 93.40 & 73.03 & 50.86 \\
Gen-3~\citep{germanidis2024gen3}            & 82.32 & 84.11 & 75.17 & 63.34 & 87.81 & 53.64 & 96.40 & 65.09 & 54.57 \\
Veo3~\citep{veo3techreport}            & 85.06 & 85.70 & 82.49 & 63.81 & 93.89 & 82.20 & 99.40 & 84.26 & 57.43 \\
\midrule
% \methodname{} & 3B & 74.87 & 76.71 & 67.53 & 52.29 & 69.48 & 39.02 & 93.40 & 41.31 & 44.65 \\
\rowcolor{gray!15} \multicolumn{10}{l}{\textit{Diffusion models}} \\
OpenSora-v1.1~\citep{zheng2024open}     & 75.66 & 77.74 & 67.36 & 50.12 & 86.76 & 40.97 & 84.20 & 52.47 & 38.63 \\
%OpenSoraPlan-v1.1 & 1B  & 78.00 & 80.91 & 66.38 & 56.85 & 76.30 & 40.35 & 86.80 & 53.11 & 27.17 \\
%OpenSora-v1.2~\citep{zheng2024open}      & 79.76 & 81.35 & 73.39 & 56.85 & 82.22 & 51.83 & 91.20 & 68.56 & 42.44 \\
CogVideoX~\citep{yang2024cogvideox}       & 80.91 & 82.18 & 75.83 & 60.82 & 83.37 & 62.63 & 98.00 & 69.90 & 51.14 \\
HunyuanVideo~\citep{kong2024hunyuanvideo} & 83.24 & 85.09 & 75.82 & 60.36 & 86.10 & 68.55 & 94.40 & 68.68 & 53.88 \\
Wan2.1-T2V~\citep{wan2025wan}         & 83.69 & 85.59 & 76.11 & 66.07 & 86.28 & 69.58 & 95.40 & 75.39 & 45.75 \\
% Hunyuan          & -   & 83.43 & 85.07 & 76.88 & 60.28 & 83.48 & 66.71 & 94.40 & 72.13 & 54.46 \\
\midrule
\rowcolor{gray!15} \multicolumn{10}{l}{\textit{Autoregressive (Diffusion) models}} \\
CogVideo~\citep{hong2022cogvideo}     & 67.01 & 72.06 & 46.83 & 38.18 & 73.40 & 18.11 & 78.20 & 18.24 & 28.24 \\
Emu3~\citep{wang2024emu3}            & 80.96 & 84.09 & 68.43 & 59.64 & 86.17 & 44.64 & 77.71 & 68.73 & 37.11 \\
NOVA~\citep{deng2024autoregressive}   & 80.12 & 80.39 & 79.05 & 59.42 & 92.00 & 77.52 & 95.20 & 77.52 & 54.06 \\
SkyReel-v2~\citep{chen2025skyreels} & 83.90 & 84.70 & 80.80 & - & - & - & - & - & - \\
MAGI-1-distill~\citep{teng2025magi}  & 77.92 & 80.98 & 65.68 & 62.43 & 82.37 & 35.08 & 84.20 & 57.75 & 26.28  \\
\midrule 
\rowcolor{gray!15} \multicolumn{10}{l}{\textit{Normalizing Flows}} \\
\methodname{} (Ours) & 78.67 & 80.24 & 72.37 & 54.48 & 86.65 & 53.48 & 94.00 & 49.84 & 47.08 \\
\methodname{}$^\dagger$ (Ours) & 79.70 & 80.76 & 75.43 & 59.73 & 80.61 & 56.04 & 98.13 & 76.08 & 48.21 \\
\methodname{}$^\dagger$ (Ours, non-Causal) & 
79.22 & 80.34 & 74.71 & 58.70 & 81.08 & 54.60 & 98.40 & 73.15 & 49.61 
\\
\bottomrule
\end{tabular}}
\caption{\textbf{Text-to-video evaluation on \textbf{VBench}~\citep{huang2024vbench}.}  
The baseline data is from the leaderboard. Following \citet{yangcogvideox}, we also evaluate with the official GPT-augmented prompts (Rewriter), with longer and more descriptive text inputs. $^\dagger$ denotes results using Rewriter prompts.
}\vspace{-10pt}
\label{tab:vbench}
\end{table*}

%% file: secs/04-exp.tex
\section{Experiments}
\subsection{Experimental Setup}
\begin{figure*}[!t]
    \centering
    \includegraphics[width=\textwidth]{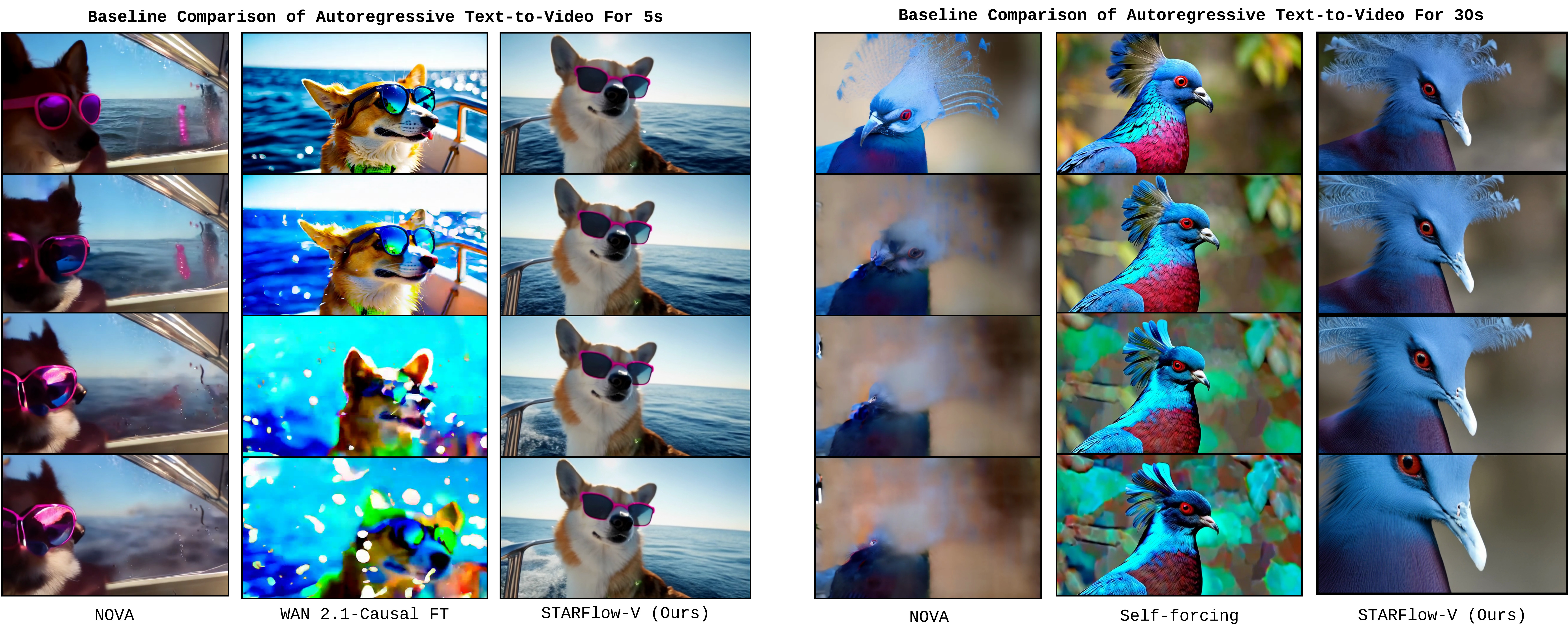}
    \caption{\methodname{} comparison against baselines on autoregressive generation for both trained length (5s) and long-horizon generation (30s). \textbf{Please refer to more video comparison in the project page.}} \vspace{-10pt}
    \label{fig:main}
\end{figure*}
\noindent\textbf{Datasets.}  
We construct a diverse and high-quality collection of video datasets to train \methodname{}. Specifically, we leverage the high-quality subset of Panda~\citep{chen2024panda} mixed with an in-house stock video dataset, with a total number of $70$M text-video pairs. For all videos, we keep their raw captions, and apply a video captioner~\citep{wang2024tarsier} to generate a longer description to cover the details. The ratio of training using raw and synthetic captions during training is $1:9$. Besides, following previous works~\citep{lin2024open}, we additionally include $400$M text-image pairs for joint training.
To support video-to-video generation and editing, we additionally finetune the pretrained \methodname{} on the Señorita~\citep{zi2025se}, a large-scale and high-quality instruction-based video editing dataset spanning 18 well-defined editing subcategories.

\noindent\textbf{Evaluation.}  
We perform both quantitative and qualitative evaluations on \methodname{}, and compare against baselines using VBench~\citep{huang2024vbench}, which benchmarks text-to-video generation across 16 dimensions, including quality, semantics, temporal consistency, and spatial reasoning.

\noindent\textbf{Model and Training Details.}
We adopt the 3D Causal VAE from WAN2.2%\footnote{\url{https://huggingface.co/Wan-AI/Wan2.2-TI2V-5B/blob/main/Wan2.2_VAE.pth}}
~\citep{wan2025wan}, which compresses spatial dimensions by $\times16$ and the temporal dimension by $\times4$ into a $48$-channel latent space. We train progressively: we initialize from an image (single-frame) model, then scale to a $7$B-parameter video model by increasing the deep-block capacity. For resolution, we use a curriculum from $384$p to $480$p while keeping the sequence length fixed at $81$ frames. For the learnable denoiser, we used a 8-layer Transformer with the same channel dimension as shallow block.
We include more implementation details in Appendix.

\noindent\textbf{Baselines.}
We compare with three baselines: (i) \textbf{WAN-2.1 Causal}, the autoregressive variant of WAN~\citep{wan2025wan} finetuned with the CausVid strategy~\citep{yin2025slow}; (ii) \textbf{Self-Forcing}~\citep{huang2025self}, finetuned from WAN-2.1 Causal-FT to mitigate train–test mismatch; and (iii) \textbf{NOVA}\citep{deng2024autoregressive}, a native autoregressive diffusion model that does not rely on vector quantization. The orginal model predicts in a chunk-based fashion. For fair comparisons, we also execute results in the pure AR settings. Besides, we also report quantitative results on VBench with official scores.

%% file: secs/04-results.tex
%\section{Results}

\subsection{Quantitative Results}

\Cref{tab:vbench} reports T2V results on VBench~\citep{huang2024vbench}. 
While \methodname{} does not yet match the strongest diffusion-based video generators, it attains performance in the same range as recent causal diffusion baselines, substantially narrowing the historical gap between NFs and diffusion models for video. 
To the best of our knowledge, \methodname{} is the \textbf{first NF-based text-to-video model} to reach this level of quality, indicating that NFs can be a viable alternative when invertibility and exact likelihood (as shown in ~\citep{zhai2024normalizing}) are desired.
We also include a variant trained without local constraints; its VBench scores remain very close to the causal version, indicating that enforcing causal structure does not incur a noticeable loss in perceptual quality.

\subsection{Qualitative Results}

\textbf{T2V \& I2V Tasks}~
As illustrated in \Cref{fig:teaser}, \methodname{} naturally supports both T2V and I2V generation.
The examples show that \methodname{} produces temporally smooth and visually faithful sequences in both settings.
Importantly, both T2V and I2V results are obtained from the \emph{same} model without additional tuning: thanks to invertibility and causal modeling, the decoder can be reused as an encoder when a conditioning image is provided.
%The first two rows show T2V results, demonstrating accurate semantic alignment with the prompt, natural motion, and consistent appearance across frames. The following two rows show I2V, where \methodname{} successfully provided reference frame into a coherent video while preserving texture, lighting, and scene composition.

% \begin{figure*}[!t]
%     \centering
%     \includegraphics[width=\textwidth]{fig/main_results1.jpeg}
%     \caption{\methodname{} examples of text and image conditioned video generation with comparison against baselines for both trained length (5s) and long-horizon generation (30s).\vspace{-10pt}} 
%     \label{fig:main}
% \end{figure*}

\begin{figure*}[!t]
    \centering
    \includegraphics[width=\linewidth]{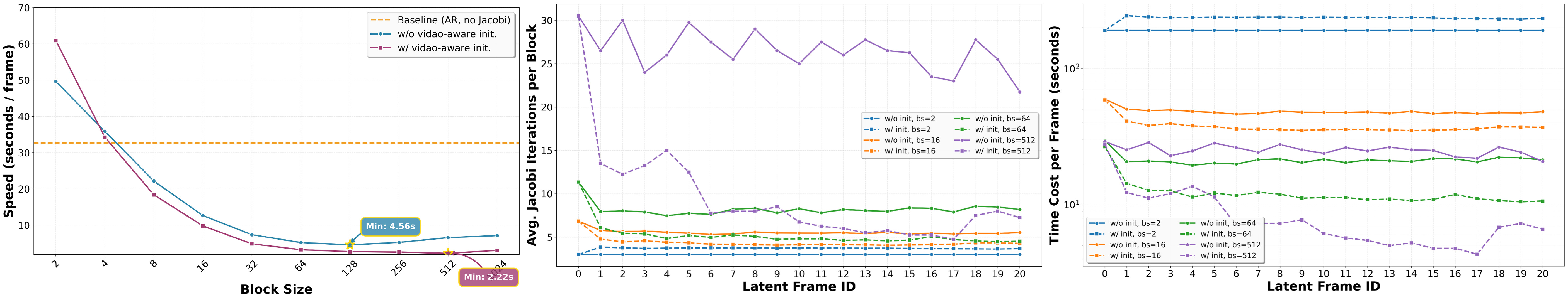}
    \caption{Comparison between speed and block size in block-wise Jacobi iteration.}
    \label{fig:gs_iter}
    \vspace{-5pt}
\end{figure*}

\noindent\textbf{V2V Tasks}~
As shown in \Cref{fig:teaser}, \methodname{} handles diverse V2V tasks from object-level to dense prediction within a single framework simply by changing the instruction. These results illustrate the  potential of using our NF-based model for general video editing and reasoning.

\noindent\textbf{Against Autoregressive Diffusion Models}~
In \Cref{fig:main}, we compare \methodname{} with two representative autoregressive diffusion models.
For the dog-with-sunglasses example, NOVA~\citep{deng2024autoregressive} exhibits gradual blurring and loss of identity, while WAN 2.1-Causal FT shows strong artifacts and color distortions.
In contrast, \methodname{} maintains clean, sharp, and temporally consistent frames, indicating stronger robustness to exposure bias.
The right block of \Cref{fig:main} further shows that \methodname{} sustains stable, coherent generations when extended to 30 seconds—well beyond its 5-second training horizon—where NOVA~\citep{deng2024autoregressive} and Self-Forcing~\citep{huang2025self} suffer from blurring, color drift, and structural deformation. We further report \textbf{quantitative} metrics for evaluating drifting effects across baselines and our model in the Appendix.

\subsection{Ablation Study}

\begin{wrapfigure}[15]{r!}{0.5\textwidth} 
    \vspace{-3mm}
    \centering
    \includegraphics[width=0.9\linewidth]{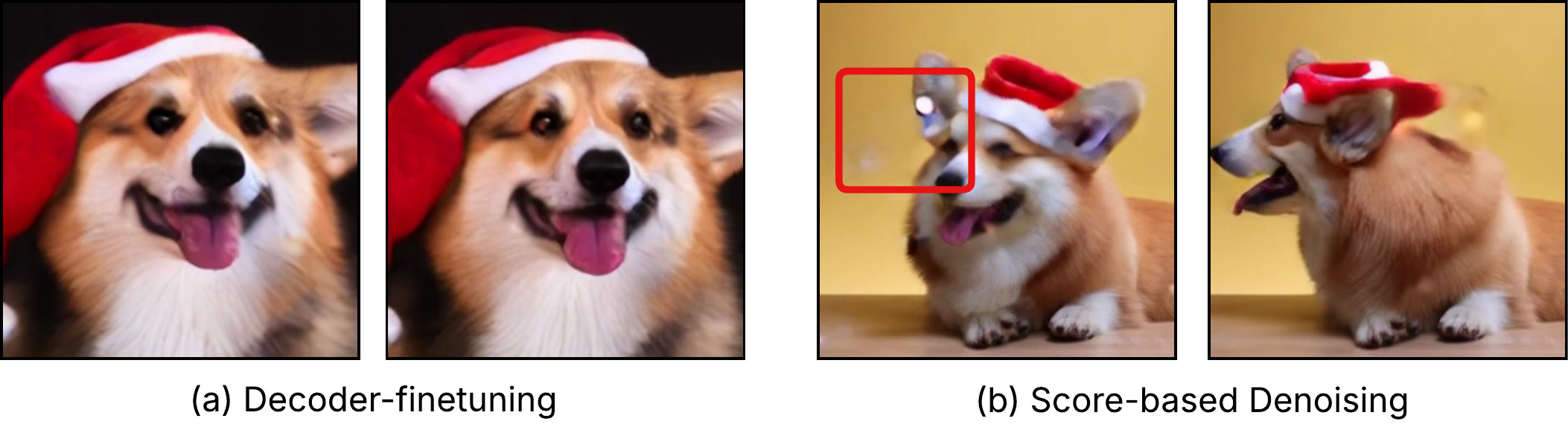}

    \medskip
    \small
    \resizebox{\linewidth}{!}{
    \begin{tabular}{lcc}
        \toprule
        Method & PSNR$\uparrow$ & SSIM$\uparrow$ \\
        \midrule
        No noise & 32.22  & 0.8907 \\
        \midrule
        Decoder fine-tuning~\citep{gu2025starflow}        & 23.95 & 0.6403 \\
        Score-based denoising~\citep{zhai2024normalizing}      & 22.05 & 0.6490 \\
        Flow-score matching (ours) & \textbf{26.69} & \textbf{0.7601} \\
        \bottomrule
    \end{tabular}}

    \caption{Ablation study for the choice of denoiser. 
    We compare video VAE reconstruction quality across denoising approaches over $1,000$ random videos with large motions.}
    \vspace{-10pt}
    \label{fig:ablation_denoiser}
\end{wrapfigure}

\noindent\textbf{Choice of Denoiser}
\Cref{fig:ablation_denoiser} provides an ablation on the denoiser design.
As shown in the top row, Decoder-finetuning~\citep{gu2025starflow} tends to lose temporal consistency with noticeable frame-to-frame jitter, while score-based denoising~\citep{zhai2024normalizing} introduces bright speckle artifacts, especially in regions of large motion. The quantitative comparison (bottom) further shows that our proposed flow–score matching achieves substantially better video reconstruction under latent-space noise injection, outperforming both alternatives by a clear margin.
% This instability stems from noisy gradients in the learned flow, motivating the need for an explicit learnable denoiser.
% This motivates the development of the flow-score matching approach. 
%Finally, as shown in (c), when the learned denoiser is constrained by a strict causal mask, the first frame denoising becomes overly smooth and blurry, which indicates the necessity of look-ahead in the network for effective temporal refinement.
%In comparison, \methodname{} shows temporally consistent and artifact-free videos.

\noindent\textbf{Hyper-parameters of Block-wise Jacobi Iteration}~
We analyze how the block size used in the block-wise Jacobi Iteration influences the runtime of the deep block. 
As shown in \Cref{fig:gs_iter} (left), the runtime initially decreases as the block size increases, reflecting better utilization of intra-block parallelism, but then rises slightly again when the block size becomes too large.
This trend suggests a trade-off: while larger block sizes increase parallelism, excessively large blocks requires more iterations within each block to achieve convergence.

We also examine the impact of video-aware initialization on runtime. 
As illustrated in \Cref{fig:gs_iter} (left), initializing the first Jacobi iteration of each frame using the converged state from the previous frame substantially reduces runtime across almost all block sizes except for small block sizes.
This improvement likely stems from the strong temporal coherence present in natural videos, where neighboring frames provide effective warm starts that appear to facilitate faster iterative updates.
Overall, video-aware initialization leads to observed improvements across block sizes.

We further analyze the runtime breakdown across latent frames in \Cref{fig:gs_iter} (right). 
Video-aware initialization yields the largest gains for large block sizes after the first frame, where convergence would otherwise require many more inner steps. 
Based on this observation, we adopt an asymmetric default strategy: \emph{use a medium block size (e.g., 64) for the first frame, and a larger block size (e.g., 512) for subsequent frames with video-aware initialization}.

% for a new frame,
% \(\vx^{(0)}_{n+1}\) is initialized from the previously converged frame \(\vx^{(k)}_{n}\).

%% file: secs/06-conclusion.tex
\section{Conclusion and Limitations}
We presented \methodname{}, an end-to-end video generative model based on autoregressive normalizing flows. 
As shown experimentally, \methodname{} delivers strong long-horizon coherence and fine-grained controllability across text-to-video, image-to-video and video-to-video tasks, and shows consistent gains over autoregressive diffusion baselines at 480p/81f. 
As a bonus, \methodname{} can be used natively for likelihood estimation.

%To accelerate sampling, we proposed inference.
While the results are encouraging, there are still limitations to overcome. 
(1) \emph{Latency.} Despite the proposed accelerated sampling, inference remains far from real time on commodity GPUs. (2) \emph{Data quality and scaling.} Progress is bounded by dataset noise and bias; we do not observe a clean scaling law under current curation.
\begin{wrapfigure}[11]{r!}{0.5\textwidth} 
    \centering
    \includegraphics[width=\linewidth]{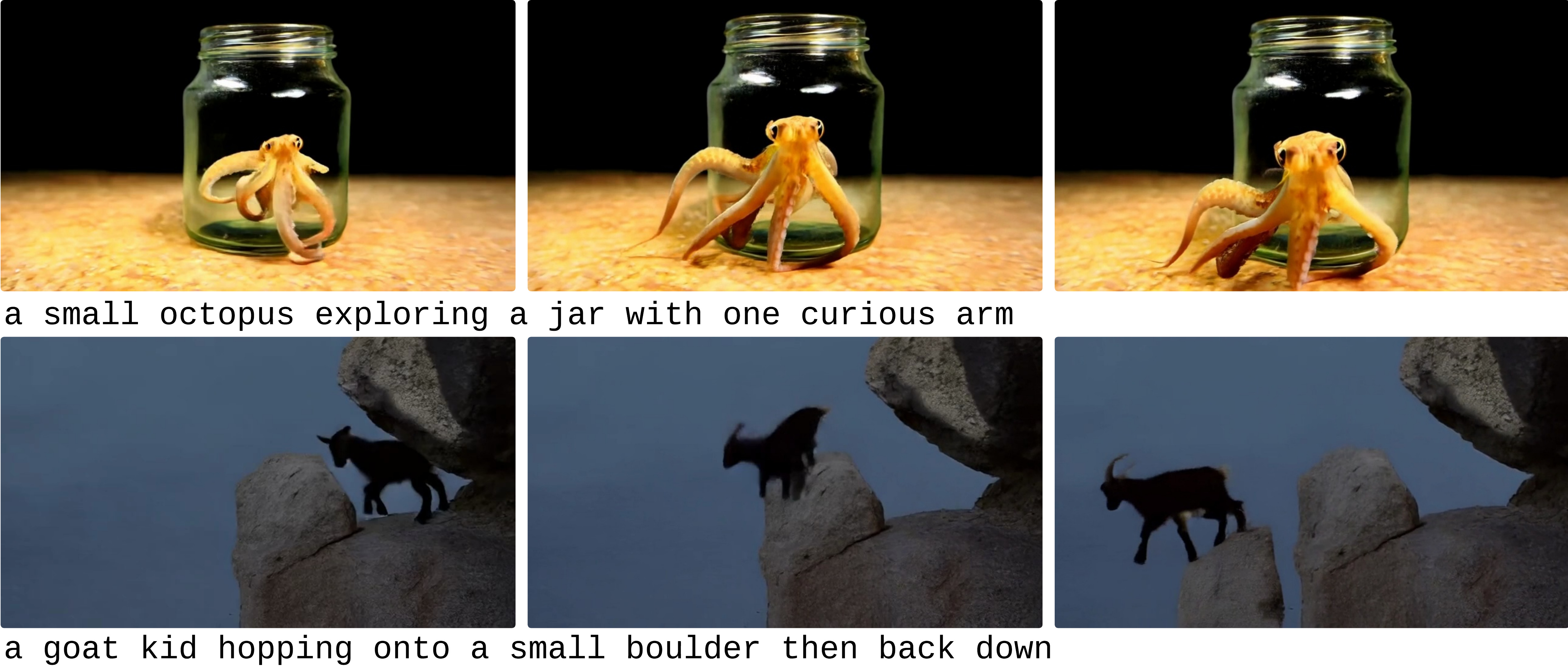}
    \caption{\label{fig.failure} Failure cases of generation from \methodname{}.}
    %\vspace{-120pt}
\end{wrapfigure}
(3) \emph{Non-physical generation.} Due to the current model scale and available data, we still observe many unrealistic, non-physical generations (see~\Cref{fig.failure}), such as an octopus passing through the wall of a jar and a rock spontaneously appearing beneath a goat just as it lands.

Looking forward, we see several promising directions. First, we aim to reduce generation latency, for example through more efficient sampling schedules and architectural optimizations. Second, we plan to study distillation and pruning to obtain compact student models that retain most of the performance of the full system. Third, we will revisit dataset curation and active data selection, with a particular focus on challenging, large-motion sequences and physically grounded scenarios; this is crucial for improving physical plausibility, reducing non-physical failure cases, and enabling clearer scaling behavior at higher fidelity.

%% file: secs/07-appx.tex
\section{Derivations and Algorithms}

\subsection{Derivation of \methodname{}.}
\label{sec:appendix-derivation}

\paragraph{(1) Why an autoregressive Gaussian model in $\vu$ is a normalizing flow.}
Let $T_\theta:\vu\mapsto\vz$ be the \emph{triangular} autoregressive map applied by the deep block $f_D$ (within a frame and across frames in the global order). For token index $i$ in that order,
\begin{equation}
\label{eq:app_ar_flow}
\vz_i \;=\; \frac{\vu_i - \mu_\theta(\vu_{<i})}{\sigma_\theta(\vu_{<i})}\,,
\qquad \sigma_\theta(\cdot)>0,
\end{equation}
with inverse
\begin{equation}
\label{eq:app_ar_inv}
\vu_i \;=\; \sigma_\theta(\vu_{<i})\, \vz_i \;+\; \mu_\theta(\vu_{<i}).
\end{equation}
Because each $\vz_i$ depends only on $(\vu_1,\ldots,\vu_i)$ and $\sigma_\theta>0$, $T_\theta$ is bijective and continuously differentiable. The Jacobian is lower triangular with diagonal entries $\partial \vz_i/\partial \vu_i = 1/\sigma_\theta(\vu_{<i})$, thus
\begin{equation}
\label{eq:app_logdet}
\log\bigl|\det J_{T_\theta}(\vu)\bigr| = - \sum_i \log \sigma_\theta(\vu_{<i}).
\end{equation}
With a standard normal prior $p_0(\vz)=\prod_i \mathcal{N}(\vz_i;0,I)$,
\begin{equation}
\label{eq:app_logpD}
\log p_D(\vu)
= \log p_0\!\bigl(T_\theta(\vu)\bigr) + \log\bigl|\det J_{T_\theta}(\vu)\bigr|
= -\tfrac12\sum_i \vz_i^2 \;-\; \sum_i \log \sigma_\theta(\vu_{<i}) \;+\; \text{const},
\end{equation}
which is essentially the regression objective through maximum likelihood estimation over $\vu$.
Therefore, the deep block realizes a valid normalizing flow. Composing with the shallow block gives $f_\theta=f_D\circ f_S$ and yields the data density in \Cref{eq.starflow_v}.

\paragraph{(2) How we get the autoregressive distribution.}
From the global–local factorization (\Cref{eq.factorize}),
\begin{equation}
\label{eq:app_frame_factor}
p_\theta(\vx)
= \prod_{n=1}^N p_D(\vu_n \mid \vu_{<n}) \;\bigl|\det J_{f_S}(\vx_n)\bigr|,\qquad \vu_n=f_S(\vx_n).
\end{equation}
Within a frame $n$, index tokens $k=1,\ldots, HW\!\cdot\!D$ in raster (or block) order and we have \Cref{eq:app_logpD} which models $p_D$ as Gaussian.
The shallow-block contributes the additional log–det $\sum_n \log|\det J_{f_S}(\vx_n)|$, forming an expressive distribution.

\paragraph{(3) Noise \& denoising: what the model looks like.}
Following the noise-augmented training (\S\ref{sec: training}), let $\tilde{\vx}=\vx+\sigma \epsilon$, $\epsilon\!\sim\!\mathcal{N}(0,I)$. The Tweedie single-step denoiser in the flow setting (\Cref{eq.score_denoise}) suggests the update $\vx \approx \tilde{\vx} + \sigma^2 \nabla_{\tilde{\vx}}\log p_\theta(\tilde{\vx})$. To avoid high-frequency artifacts and to preserve streamability, we fit a \emph{causal} denoiser $s_\phi$ via flow-score matching (\Cref{eq:fsm_loss}) and then use
\begin{equation}
\label{eq:app_tweedie_s}
\hat{\vx} \;=\; \tilde{\vx} \;+\; \sigma\, s_\phi(\tilde{\vx}) \;\;\approx\;\; \tilde{\vx} \;+\; \sigma^2 \nabla_{\tilde{\vx}}\log p_\theta(\tilde{\vx}),
\end{equation}
where $s_\phi$ uses a block-causal mask with at most one-frame look-ahead to retain strict streamability.

\subsection{Training}
\label{sec:appendix-algorithm}

\Cref{alg:training} shows the training algorithm of \methodname{}{} for both the flow and the learnable denoiser.

\begin{algorithm}[t]
\caption{Training \methodname{} with noise augmentation and flow-score matching}
\label{alg:training}
\begin{algorithmic}[1]
\Require video dataset $\mathcal{D}$; noise level $\sigma$; FSM weight $\lambda_{\text{den}}$
\Repeat
  \State Sample mini-batch $\vx\sim\mathcal{D}$ and noise $\boldsymbol{\epsilon}\sim\mathcal{N}(0,I)$
  \State \textbf{Noise-augment:} $\tilde{\vx} \gets \vx + \sigma\,\boldsymbol{\epsilon}$ \Comment{as in \S\ref{sec: training}}
  \State \textbf{Shallow forward:} $\vu \gets f_S(\tilde{\vx})$ \Comment{alternating masked AF blocks, within-frame}
  \State \textbf{Deep forward:} $\vz \gets f_D(\vu)$ \Comment{causal Transformer AF over global order}
  \State \textbf{Standard NF NLL:} $\mathcal{L}_{\text{NLL}}(\theta) \gets -\big[\log p_0(\vz)
         + \log|\det J_{f_D}(\vu)||\det J_{f_S}(\tilde{\vx})|\big]$ 
  \State \textbf{Score target (stop-grad):} 
         $\mathbf{t} \gets \sigma\,\nabla_{\tilde{\vx}}\log p_\theta(\tilde{\vx})$ \Comment{reuse backward pass of $\mathcal{L}_{\text{NLL}}$; detach}
  \State \textbf{Flow-score Matching:} $\mathcal{L}_{\text{FSM}}(\phi) \gets \|\; s_\phi(\tilde{\vx}) - \mathbf{t} \;\|_2^2$
  \State \textbf{Total loss:} $\mathcal{L} \gets \mathcal{L}_{\text{NLL}}(\theta) + \lambda_{\text{den}}\,\mathcal{L}_{\text{FSM}}(\phi)$
  \State \textbf{Update:} $(\theta,\phi) \leftarrow (\theta,\phi) - \eta\,\nabla \mathcal{L}$
\Until{convergence}
\end{algorithmic}
\end{algorithm}

\subsection{Inference}
\label{sec:appendix-inference}

\begin{algorithm}[t]
\caption{Autoregressive sampling (\(\vz\!\to\!\vu\!\to\!\vx\))}
\label{alg:ar-sampling}
\begin{algorithmic}[1]
\Require length $N$ (frames or tokens), base prior $p_0(\vz)=\mathcal{N}(0,I)$, shallow inverse $f_S^{-1}$, deep inverse $f_D^{-1}$, token order $\prec$
\State Sample $\vz \sim \mathcal{N}(0,I)$ with the target shape
\State Initialize an empty latent sequence $\vu$
\For{each element $i$ in global order $\prec$} \Comment{causal AR over frames and within-frame tokens}
  \State Compute $(\mu_i,\sigma_i)$: $(\mu_i,\sigma_i)\!\leftarrow\! f_D\big(\vu_{<i}\big)$
  \State Invert deep at position $i$: $\vu_i \leftarrow \sigma_i\, \vz_i + \mu_i$ \Comment{$f_D^{-1}$, triangular}
\EndFor
\State Invert shallow block: $\vx \leftarrow f_S^{-1}(\vu)$
\State \textbf{(One-step corrector)} $\vx \leftarrow \vx + \sigma_{\text{test}}\, s_\phi(\vx)$
\State \Return $\vx$
\end{algorithmic}
\end{algorithm}

\begin{algorithm}[t]
\caption{Jacobi-style parallel inversion of the deep autoregressive block}
\label{alg:jacobi}
\begin{algorithmic}[1]
\Require base latent $\vz$; 
initial guess $\vu^{(0)}$ (e.g., zeros); 
block partition $\mathcal{B}=\{B_1,\dots,B_J\}$ (non-overlapping, block-causal,  $|B_j| = 4|B_1|$ for all block $j > 1$); max iters $T$; Frame size $F$; tolerance $\tau$
% \For{$t=0,1,2,\ldots,T-1$}
  % \ForAll{blocks $B \in \mathcal{B}$ \textbf{in parallel}}
  \For{$j=1,2,\ldots,J$}
    \State $[a,b] \leftarrow B_j$ \Comment{indices of the $j$-th block}
    \If{$j=1$ \textbf{and} $a > F$}
        \State Initialize $\vu_{a:b} \leftarrow \vu_{a:b}^{(0)}$ \Comment{random initialization}
    \Else
        \State Initialize $\vu_{a:b} \leftarrow \vu_{a-F:b-F}$ \Comment{initialization from past frame}
    \EndIf
    \Repeat
        \State $t \leftarrow t+1$
         \ForAll{$i \in B_j$ \textbf{in parallel}}
            \State $(\mu_i^{(t)},\sigma_i^{(t)}) \leftarrow f_D\big(\vu^{(t)}_{<i}\big)$
            \State $\vu_i^{(t+1)} \leftarrow \sigma_i^{(t)}\, \vz_i + \mu_i^{(t)}$
         \EndFor
         \Until{$\frac{\|\vu^{(t+1)}-\vu^{(t)}\|_2}{\|\vu^{(t)}\|_2+\varepsilon} \le \tau$ \textbf{or} $t=T$}
    \State $\vu_{a:b} \leftarrow \vu^{(t)}_{a:b}$
  \EndFor
% \EndFor
\State \textbf{Shallow inverse:} $\vx \leftarrow f_S^{-1}\!\big(\vu^{(t+1)}\big)$
\State \textbf{(One-step corrector)} $\vx \leftarrow \vx + \sigma_{\text{test}}\, s_\phi(\vx)$
\State \Return $\vx$
\end{algorithmic}
\end{algorithm}

\paragraph{Remarks.}
(i) When the deep map is sufficiently contractive in $\vu$ (e.g., via scale clamping), the Jacobi iteration converges rapidly and enables wide parallelism within each block $B$. 
(ii) A common choice for $\mathcal{B}$ is to use spatial tiles per frame (no intra-tile dependencies) or even/odd raster groups, preserving the block-causal mask used in training.

\begin{algorithm}[t]
\caption{Streaming long-sequence generation via \emph{re-encode with forward}}
\label{alg:streaming-reencode}
\begin{algorithmic}[1]
\Require target length $T$ (frames), window size $W$ ($W\!\ll\!T$); deep inverse $f_D^{-1}$; shallow inverse $f_S^{-1}$; shallow forward $f_S$; deep forward $f_D$; prior $p_0(\vz)$
\State Initialize caches $\mathsf{KV}\leftarrow \varnothing$, latent buffer $\mathsf{U}\leftarrow \varnothing$
\For{$t=1$ to $T$}
   \State \textbf{Sample base:} $\vz_t \sim \mathcal{N}(0,I)$ for the next frame (or token block)
   \State \textbf{Deep inverse:} using cached state, compute $\vu_t \leftarrow f_D^{-1}(\vz_t \,;\, \mathsf{KV})$ and update the $\mathsf{KV}$ cache.
   \State \textbf{Shallow inverse:} $\vx_t \leftarrow f_S^{-1}(\vu_t)$
   \State \textbf{Emit} $\vx_t$
   \State \textbf{Re-encode (forward):} $\hat{\vu}_t \leftarrow f_S(\vx_t)$ \Comment{brings the produced frame back to $U$-space}
   \State \textbf{Update deep state:} run $f_D$ \emph{forward} on $\hat{\vu}_t$ to refresh $\mathsf{KV}$ (no sampling): $\_ \leftarrow f_D(\hat{\vu}_t; \mathsf{KV})$
   \State \textbf{Maintain sliding window:} push $\hat{\vu}_t$ into buffer $\mathsf{U}$; if $|\mathsf{U}|>W$ pop the oldest
\EndFor
\State \Return $\{\vx_t\}_{t=1}^T$
\end{algorithmic}
\end{algorithm}

\section{Implementation Details}

\subsection{Architecture Design}
\label{sec:appendix-arch}

\begin{table}[h]
\centering
%\small
\setlength{\tabcolsep}{3pt}
\begin{tabular}{@{}lcc@{}}
\toprule
 & \textbf{3B} & \textbf{7B} \\
\midrule
Params & $\sim$3B & $\sim$7B \\
$f_D$ width & 3072 & 4096 \\
$f_S$ & \multicolumn{2}{c}{identical (alt. masked AF; width $d_S$, depth $L_S$)} \\
Denoiser $s_\phi$ & \multicolumn{2}{c}{8-layer Transformer, block-causal mask} \\
Init & from scratch & finetune from 3B \\
\bottomrule
\end{tabular}
\caption{Minimal comparison. Only $f_D$ width differs; $f_S$ and $s_\phi$ are unchanged.}
\label{tab:arch}
\end{table}

% --- Guaranteed-fit fallback (uncomment if still too wide) ---
% \begin{table}[h]
% \centering
% \resizebox{\columnwidth}{!}{%
% \begin{tabular}{@{}lcc@{}}
% \toprule
%  & \textbf{3B} & \textbf{7B} \\
% \midrule
% Params & $\sim$3B & $\sim$7B \\
% $f_D$ width & 3072 & 4096 \\
% $f_S$ & \multicolumn{2}{c}{identical (alt. masked AF; width $d_S$, depth $L_S$)} \\
% Denoiser $s_\phi$ & \multicolumn{2}{c}{8-layer Transformer, block-causal mask} \\
% Init & from scratch & finetune from 3B \\
% \bottomrule
% \end{tabular}}
% \caption{Minimal comparison. Only $f_D$ width differs; $f_S$ and $s_\phi$ are unchanged.}
% \label{tab:arch-tight}
% \end{table}

\noindent\textbf{3B.} Same size as STARFlow but for \emph{video}. The deep block $f_D$ uses width $3072$ (depth $L_D$, heads $H_D$). The shallow stack $f_S$ (alternating masked affine flows) and the denoiser $s_\phi$ (8-layer Transformer with block-causal mask) follow the standard design.

\noindent\textbf{7B.} Initialized from the 3B checkpoint and \emph{only} widens the deep block $f_D$ channels from $3072$ to $4096$. The shallow stack $f_S$ and denoiser $s_\phi$ remain identical (same depths, heads, and widths).

\subsection{Training Details}

\methodname{} is trained on $96$ H100 GPUs using approximately 20 million videos. In all the experiments, we share the following training configuration for our proposed \methodname{}. 
\begin{verbatim}
training config:
    batch_size=96
    optimizer='AdamW'
    adam_beta1=0.9
    adam_beta2=0.95
    adam_eps=1e-8
    learning_rate=5e-5
    min_learning_rate=1e-6
    learning_rate_schedule=cosine
    weight_decay=1e-4
    mixed_precision_training=bf16
\end{verbatim}

\paragraph{Progressive Video Training} We adopt a progressive multi-stage training paradigm that gradually increases model size, resolution, and temporal horizon for stable and effective optimization.
\begin{itemize}
    \item \textbf{3B Text-to-Image Training:} We initialize a 3B text-to-image model from the pretrained StarFlow~\citep{gu2025starflow}, establishing a strong visual–textual backbone before introducing temporal modeling.
    \item \textbf{3B Image-Video Joint Training (384P, 45 frames):} The 3B model is then jointly trained on low-resolution images and videos at 384P. Each training clip contains 45 frames sampled at 16 fps, enabling the model to acquire short-term temporal dynamics.
    \item \textbf{7B Image-Video Joint Training (384P, 81 frames):} We expand the model to 7B parameters and continue joint training at 384P, doubling the temporal horizon from 45 to 81 frames to strengthen long-range temporal reasoning.
    \item \textbf{7B Image-Video Joint Training (480P, 81 frames):} Finally, we train the 7B model on higher-resolution 480P images and videos while maintaining the 81-frame temporal window.
\end{itemize}

\paragraph{Mixed-Resolution Training} 
% \methodname{} supports \textit{mixted resolutions}, preserving each frame's native aspect ratio.

\methodname{} is designed to support \textit{mixed-resolution} inputs, allowing each frame to retain its native aspect ratio and spatial resolution. 
Similar to \citet{gu2025starflow}, we assign each video sequence to one of nine predefined aspect-ratio bins, since all frames within a video share the same ratio.
The pre-defined bins are 21:9, 16:9, 3:2, 5:4, 1:1, 4:5, 2:3, 9:16, and 9:21.
To make the model explicitly aware of these visual formats, we incorporate both the fps and aspect-ratio tag into the text caption:
\begin{verbatim}
    A video with {fps} fps:
    {original_caption}
    in a {aspect_ratio} aspect ratio.
\end{verbatim}

\paragraph{Gradient Control}
We monitor the gradient norm throughout training to ensure stability. 
Specifically, to prevent gradient explosion, we enable gradient skipping after the first 100 steps: if the gradient norm exceeds a threshold of 1, the update for that step is skipped.
This adaptive strategy stabilizes early training while maintaining convergence efficiency later on. 
\subsection{Baseline Details}

\paragraph{WAN-2.1 Causal-FT} is the autoregressive variant of WAN~\citep{wan2025wan}.
Specifically, we adopt Wan2.1-T2V-1.3B, a Flow Matching–based model, as the base model.
Following the CausVid initialization strategy~\citep{yin2025slow}, the base model is fine-tuned with causal attention masking on 16k ODE solution pairs generated from the model itself. 
In practice, we leverage the ODE initialization checkpoint released with the official Self-Forcing~\citep{huang2025self} repository, which corresponds exactly to the configuration of our WAN-2.1 Causal-FT setup.
% —that is, the Wan2.1-T2V-1.3B model fine-tuned with causal attention on 16k ODE solution pairs as described above.

\paragraph{NOVA AR~\citep{deng2024autoregressive}} is an autoregressive video generator that does not rely on vector quantization.
It reformulates video generation as non-quantized autoregressive modeling that performs temporal frame-by-frame prediction while generating spatial token sets within each frame in a flexible, set-by-set manner.
To support autoregressive modeling with continuous tokens, NOVA leverages a lightweight diffusion head that models the distribution of each continuous token~\citep{li2024autoregressive}.
In this work, we directly compare the pure AR version of NOVA, where the model predicts each latent frame with diffusion for a fair comparison.
\input{tabs/drifting}

\begin{figure}[t]
    \centering
    \includegraphics[width=\linewidth]{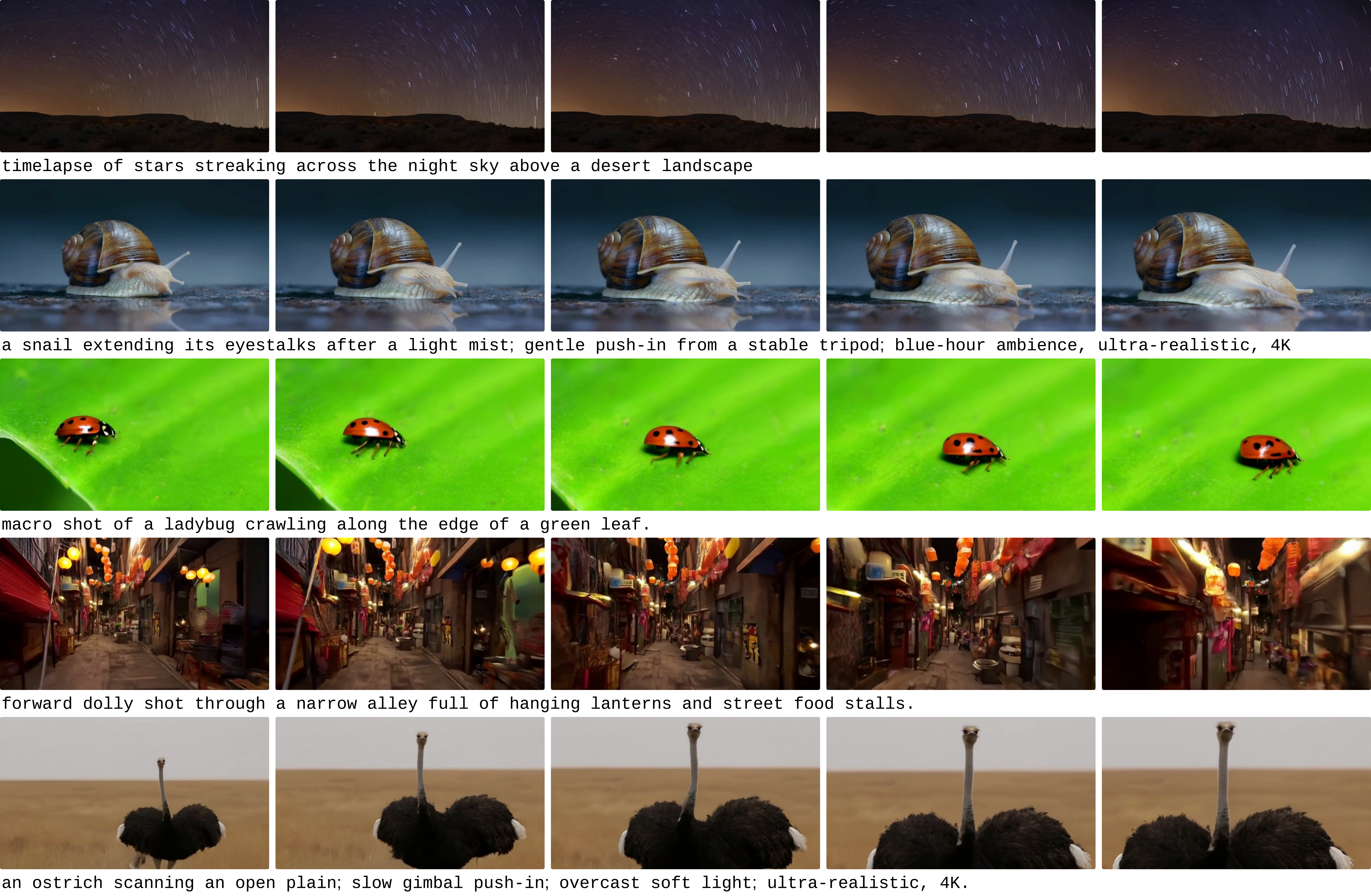}
    \caption{Generated samples from \methodname{} given text prompts. All videos are at 480p 16fps and 5s.}
    \label{fig:additional_t2v}
\end{figure}
\begin{figure}[t]
    \centering
    \includegraphics[width=\linewidth]{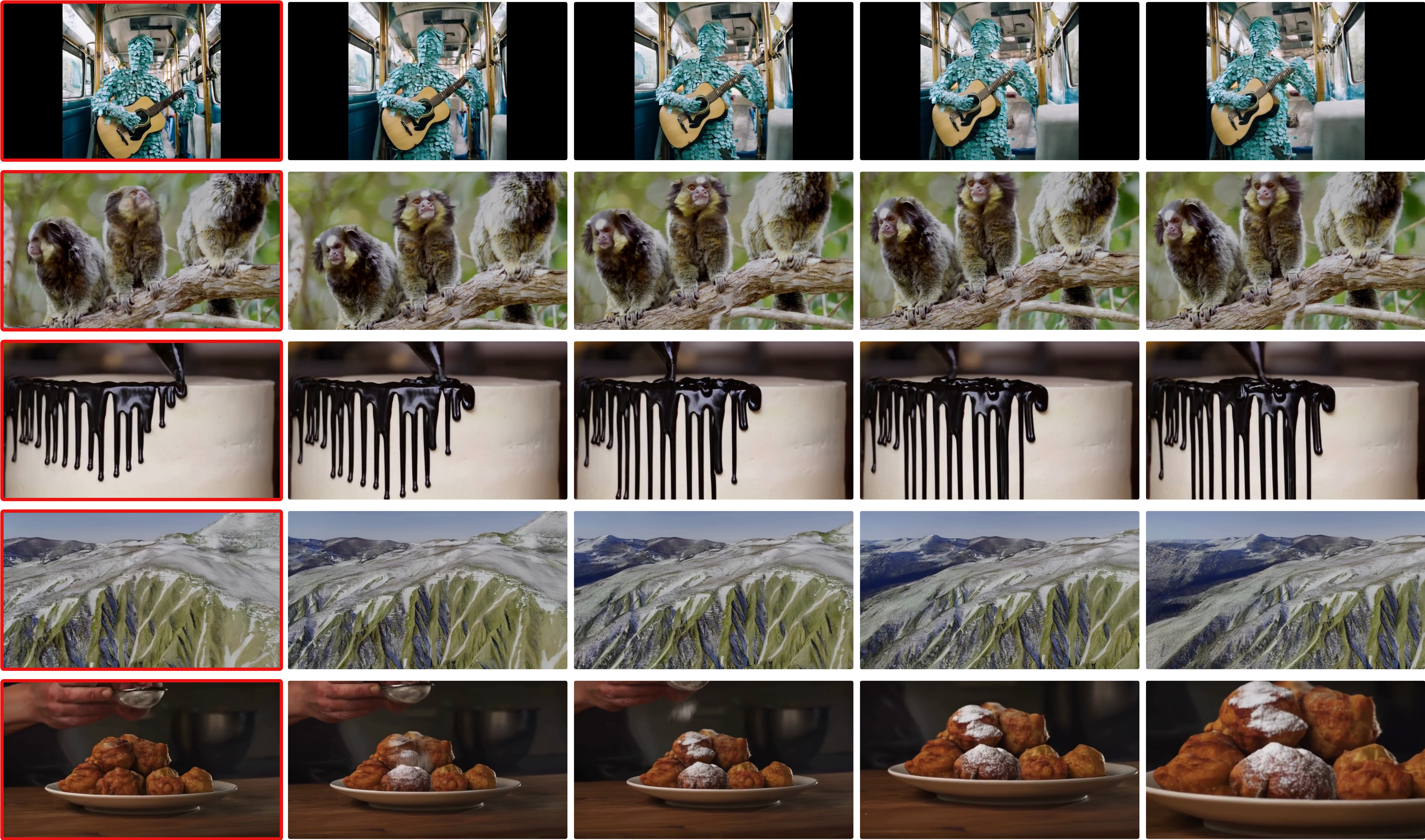}
    \caption{Generated samples from \methodname{} given the first frame. All videos are at 480p 16fps and 5s.}
    \label{fig:additional_i2v}
\end{figure}
\begin{figure}[t]
    \centering
    \includegraphics[width=\linewidth]{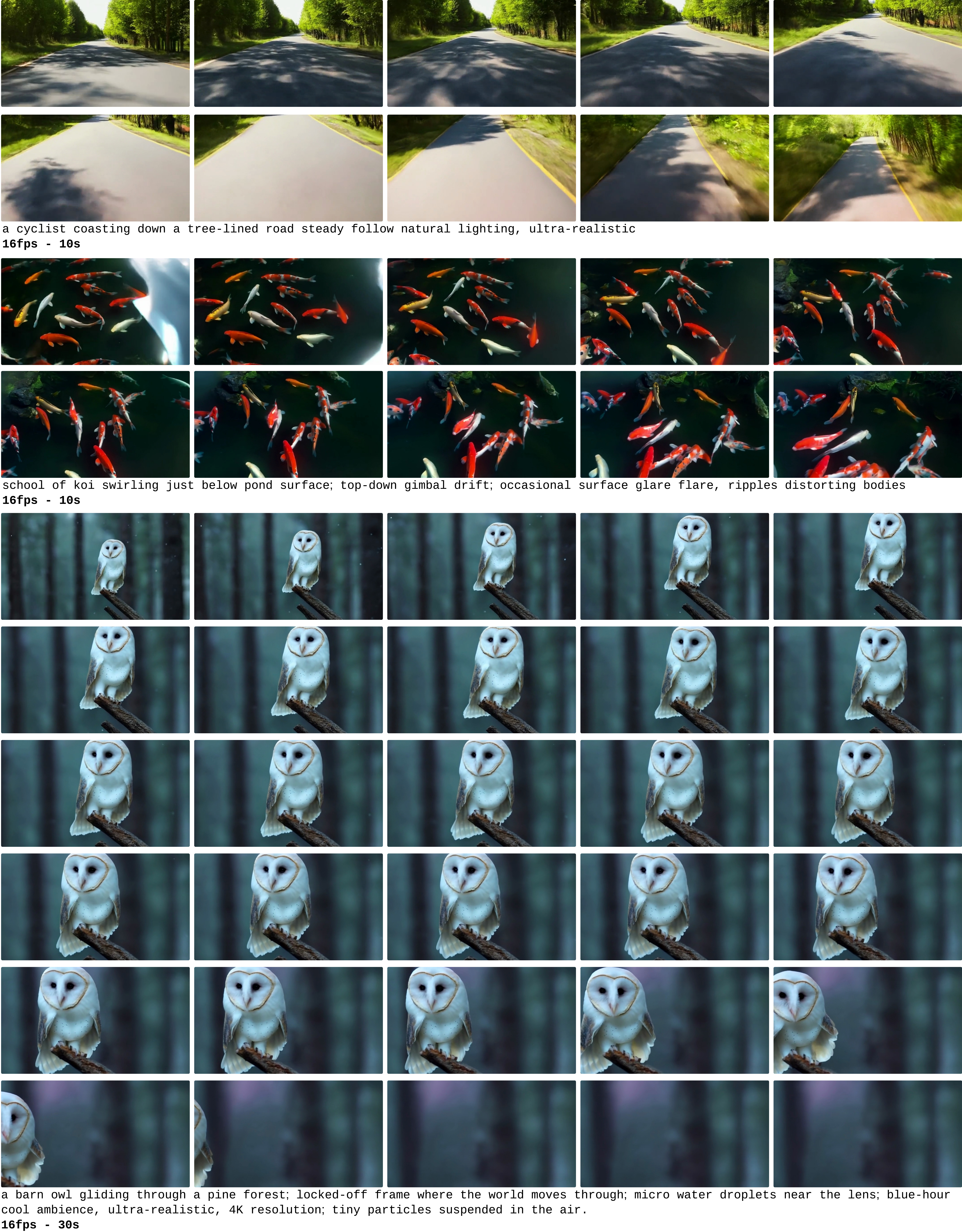}
    \caption{Generated samples from \methodname{} given text prompts and extended with overlapping frames. For each segment, we generate 21 latent frames with 4 latent frames in overlap. Both videos are at 480p 16fps.}
    \label{fig:additional_long}
\end{figure}
\section{Additional Experimental Details and Results}

\subsection{Quantitative Comparison with Autoregressive Diffusion baselines}
To evaluate the robustness of video generation under autoregressive generation, we compare \methodname{} with autoregressive diffusion models, including NOVA AR~\citep{deng2024autoregressive} and WAN 2.1-Causal FT. Here, NOVA AR refers to the fully autoregressive video generation variant which is different from the reported in the official paper.
\Cref{tab:drift} compares these models across a diverse set of evaluation dimensions defined in VBench~\citep{huang2024vbench}.
As shown in \Cref{tab:drift}, \methodname{} substantially outperforms the autoregressive diffusion baselines across all dimensions.
Both NOVA AR and WAN~2.1-Causal FT exhibit clear signs of autoregressive degradation in their generated videos. Specifically, NOVA AR suffers from pronounced error accumulation, leading to increasing blur and content collapse as the video progresses. 
And WAN~2.1-Causal FT produces noticeable temporal inconsistency and flickering throughout the video. 
These failure modes are reflected in their lower scores, underscoring the difficulty of maintaining robustness in autoregressive video generation. And it further highlights the strength of our approach.

\subsection{Video-to-Video Generation}

To support video-to-video generation and editing, we additionally finetune the pretrained \methodname{} (7B, 384P, 81 frames) on the Señorita~\citep{zi2025se}, a large-scale and high-quality instruction-based video editing dataset spanning 18 well-defined editing subcategories.
Each training sample in Señorita consists of a 33-frame input video paired with a 33-frame edited target video. The model is also trained on videos with 16fps. This finetuning stage equips \methodname{} with precise editing capabilities while preserving temporal coherence and motion consistency.
During finetuning, we concatenate the input and target videos along the temporal dimension to form a single training sequence.

% For additional qualitative results and video demonstrations, please refer to the HTML viewer included in the supplementary materials.

\subsection{Additional Samples}
We show additional samples at \Cref{fig:additional_t2v,fig:additional_i2v,fig:additional_long}. Besides, we provide more video generation comparison in our official codebase at \url{https://github.com/apple/ml-starflow}.

%% file: tabs/drifting.tex
\begin{table*}[!bhtp]
\centering
\resizebox{0.95\linewidth}{!}{
\begin{tabular}{lcccccccc}
\toprule
Model  & Total & Quality & Semantic & Aesthetic & Object & Human & Spatial & Scene \\
\midrule
\rowcolor{gray!15} \multicolumn{9}{l}{\textit{Autoregressive (Diffusion) models}} \\
NOVA AR$\dagger$~\citep{deng2024autoregressive}    
& 75.31 & 77.46 & 66.70 
& 56.04 & 79.68 & 94.20 & 66.07 & 47.83 \\

WAN 2.1-Causal FT$\dagger$ 
& 74.96 & 77.41 & 65.15 
& 56.04 & 76.51 & 94.20 & 53.25 & 47.83 \\

\midrule 
\rowcolor{gray!15} \multicolumn{9}{l}{\textit{Normalizing Flows}} \\
\methodname{}$\dagger$ (Ours) 
& \textbf{79.70} & \textbf{80.76} & \textbf{75.43} 
& \textbf{59.73} & \textbf{80.61} & \textbf{98.13} & \textbf{76.08} & \textbf{48.21} \\
\bottomrule
\end{tabular}}
\caption{\textbf{Performance comparison of autoregressive video generation models on VBench~\citep{huang2024vbench}.} Following \citet{yangcogvideox}, we evaluate with the official GPT-augmented prompts (noted as $\dagger$) 
}\vspace{-10pt}
\label{tab:drift}
\end{table*}